# A Data-Driven Approach for Linear and Nonlinear Damage Detection Using Variational Mode Decomposition and GARCH Model

**Vahid Reza Gharehbaghi[1], Hashem Kalbkhani[2], Ehsan Noroozinejad Farsangi[3], T.Y. Yang[4], Seyedali Mirjalili[5,6]**

*[1] Adjunct Associate Research Fellow, School of Civil Engineering and Surveying, University of Southern Queensland, Springfield Campus, Queensland, Australia*
*[2] A/Professor, Faculty of Electrical Engineering, Urmia University of Technology, Urmia, Iran*
*[3]A/Professor, Faculty of Civil and Surveying Engineering, Graduate University of Advanced Technology, Kerman, Iran*
*[4]Professor, Department of Civil Engineering, The University of British Columbia, Vancouver, Canada*
*[5]A/Professor, Centre for Artificial Intelligence Research and Optimisation, Torrens University Australia, Brisbane, Australia*
*[6]Yonsei Frontier Lab, Yonsei University, Seoul, Korea*

## Abstract

In this article, an original data-driven approach is proposed to detect both linear and nonlinear damages in structures using output-only responses. The method deploys variational mode decomposition (VMD) and generalized autoregressive conditional heteroscedasticity (GARCH) model for signal processing and feature extraction. To this end, VMD decomposes the response-signals are first decomposed to intrinsic mode functions (IMFs), and then, GARCH model is utilized to represent the statistics of IMFs. The model coefficients' of IMFs construct the primary feature vector. Kernel-based principal component analysis (PCA) and linear discriminant analysis (LDA) are utilized to reduce the redundancy from the primary features by mapping them to the new feature space. The informative features are then fed separately into three supervised classifiers: support vector machine (SVM), k-nearest neighbor (kNN), and fine tree. The performance of the proposed method is evaluated on two experimental scaled models in terms of linear and nonlinear damage assessment. Kurtosis and ARCH tests proved the compatibility of GARCH model. The results demonstrate that the proposed technique reaches the accuracy of 100% and 98.82% in classifying linear and nonlinear damages, respectively. Also, its accuracy is higher than 80% in the presence of noise with a signal-to-noise ratio (SNR) higher than 10 dB.

# I. Introduction

Today's current structural engineering industry requires consideration to be directed towards structural health monitoring (SHM) and optimizing safety. With forecasts of increasing worlds' population, structural infrastructure shall be subject to increased loading and deformation. To decrease the effects and consequences of structural deterioration, SHM processes are required more frequently, with high levels of accuracy necessary to achieve asset preservation. Hence, there has been a surge in interest surrounding SHM and the development of automated defect evaluation systems in an attempt to maintain existing structural networks and allow for asset expansion.

Concerning structural behavior, damage leads to deviations in the structure's dynamic characteristics and is considered a reliable indication of anomaly diagnosis. Also, it might cause a system with a typically linear behavior to demonstrate nonlinear responses, including cracking, impacts and rattling, delamination, stick or slip, rub, or deformation in connections [1, 2]. Nonlinear behavior is supposed to be unpredictable and more sophisticated compared to the linear one. As a case in point, it has been proven through experimental investigation that natural frequencies could rise instead of decrease on breathing phenomena [3]. This reaction originates from the fact that the crack conversely opens and closes in the experimental test. Subsequently, the detection of nonlinear anomalies is considered more challenging compared to linear damages [4].

Over decades, researchers proposed several techniques in terms of anomaly identification. Generally speaking, such methods are divided into physics-based (or model-based) and data-driven approaches [5]. In the physics-based, anomalies are tracked utilizing monitoring variations within the simulated responses from the structural numerical model [6]. This model is a detailed mathematical abstraction linking a studied system's input and output variables employing known or presumed properties [7]. Post analysis is demanded for determining damage location and qualification. Finite element methods (FEMs), boundary element methods (BEMs), and spectral finite element methods (SFEMs) are some of the techniques used in this regard. However, FEMs are considered the systematic method compared to the others due to their compliance in modeling complicated structures [8]. In the occurrence of damage, particular parameters of the simulated models are updated according to response measurements. Optimization algorithms are typically used to minimize variations between experimental and numerical responses by comparing mechanical characteristics of stiffness, damping, or mass [6].

Despite the broad potential of physics-based approaches in damage assessment, especially for the evolution of complex systems such as multi-stories buildings and multi-span bridges, they have some limitations. For example, exact modeling of a structure entails sufficient information regarding different components of a monitored system, such as loading states, boundary conditions, material properties, and precise coordinates of members. Moreover, optimization solutions commonly experience numerical instability as well as ill-conditions dilemma [9]. The performance of such optimization techniques substantially degrades proportionally to the number of variables in the problem.

On the other side, data-driven SHM provides bottom-up solutions founded on tracking changes within the output signals appropriate for complex systems where the knowledge about geometries, properties, and initial conditions is limited [5]. Any sudden changes in the output signals are observed and analyzed through signal processing tools and pattern recognition procedures to determine probable damage. Independence for having an initial model and prior knowledge causes data-driven SHM to be a faster technique and an economical and practical solution for online SHM. Signal processing techniques synthesize, modify, analyze the recorded responses, and highlight different features in time, frequency, and frequency domains. Machine Learning algorithms are typically employed to identify and interpret features extracted from signals and recognize generated patterns in conjunction with such methods. Machine learning includes clustering, regression, neural networks, ensemble learning, deep learning, Bayesian methods, instance-based, decision trees, and dimensionality reduction [10].

Data-driven methods are helpful compared to physics-based techniques when [11], first, the structure's physical characteristics are unavailable or challenging to be modeled. Secondly, there are an adequate amount of sensors installed for capturing the structure's responses. Thirdly, the computational operations are costly in the SHM project; in addition, multi-physics models consist of more physical processes in a system (e.g., thermal interactions, water precipitation, magnetostatic and chemical reactions) may not seem efficient for utilizing a large amount of sensor data. The accuracy of physics-based depends on the response measurements; the best performance is achieved in an environment with the slightest noise. In real-world structures and especially for in-servicing conditions, however, the amount of noise is considerable. As such, data-driven damage identifications deploying actual responses have revealed preferable adaptability and thereby turned into an inspiring solution in the realm of SHM [10].

## II. Need For Research

Although nonlinear damage has been studied before and practical solutions are proposed in this realm, most focus on damage identification as the first level based on Rytter's classification levels in SHM [12]. Hence, limited research has been conducted to reach higher levels (e.g., damage localization and classification). This study attempts to address nonlinear damage detection in building structures through a robust data-driven approach. Adverse conditions such as environmental and operational effects in recording responses and analyzing signals are the other crucial points that should be considered. These issues become more though in the case of buildings where the story correlations can affect the structural responses. Therefore, proposing a robust model with appropriate precision in identifying different kinds of linear and nonlinear anomalies considering these issues leads to a practical approach in assessing real-world structures under adverse conditions.

Accordingly, the rest of the paper is organized as follows. In Section 2, related works are discussed, and gaps are highlighted once again. Case studies are presented in detail in Section 3. Section 4 provides the details of the proposed data-driven approach. Experimental results and discussion are given in Section 5. Finally, Section 6 concludes the work and suggests future directions.

## III. Background

Signal processing techniques play a fundamental role in data-driven SHM for analysis responses in time, frequency, or time-frequency domains. Fourier spectra, spectrum analysis, difference frequency analysis, and the high-frequency resonance technique are appropriate for damage identification, especially for gear faults and roller bearings [13]. Wavelets proved the efficiency for damage and deterioration detection in building structures based on a stochastic approach [14]. Fourier transform (FT) and fast Fourier transform (FFT) are considered the main concepts for anomaly detection. A time series model is a promising tool for simulating and predicting structural signals in the time domain.  Since this method is based on a partial structural dynamics model, tit can identify even a small number of vibrations [15].  In this area, autoregressive (AR) models are investigated for damage and deterioration detection in buildings and bridges [16-18]. Auto-regressive and moving average model (ARMA), as well as generalized autoregressive conditional heteroscedasticity model (GARCH), have proved to be beneficial for nonlinear damage identification in building specimens [19]. Transient behaviors caused by damage or adverse environmental conditions can be recognized through a signal's time-frequency form [20].

In a broad perspective, the real-world signals are linear and stationary and are coupled with noise. Consequently, linear signal processing techniques, such as spectral analysis, are not appropriate in this realm of scope [21]. Hilbert–Huang transform (HHT), introduced by Huang et al. [22], consists of two sequential steps. The first step, called empirical mode decomposition (EMD), separates the complicated initial signal into a determined and commonly limited number of intrinsic mode functions (IMFs) or modes. Each mode is an oscillatory function with time-varying frequencies that reveal the input signals' local features and correspond to different frequencies and a residue [23, 24]. The algorithm detects the maxima/minima recursively, assesses the envelopes using the extrema, and removes the average envelopes, which leads to isolating high-frequency bands.[25]. In the next step, the Hilbert transform (HT) includes each IMF's orthogonal pair with 90 degrees difference in the phase [26]. As a result, each IMF set and the corresponding pair can evaluate instant variations of signal magnitude and frequency concerning time. Compared to wavelet analysis and Fourier transform, EMD benefits from tracing out the IMFs by interpolating between the extremums instead of using any given wavelet basis. Despite the wide usage of EMD in a variety of time-frequency applications such as medical [27], economics [28], climate predictions [29], SHM [30, 31], and many other fields, it may dace with some issues like sensitivity to noise and sampling frequency which cause the performance relies on the frequency ratio [25, 31, 32].

Some modified algorithms have been developed, including ensemble EMD (EEMD), complete ensemble EMD with adaptive noise (CEEMDAN), and Variational mode decomposition (VMD) [32] to address these limitations. VMD is a relatively new algorithm that decomposes a signal into distinctive amplitude and frequency adjusted sub-signals where together they reproduce the primary input signal [32]. This approach is entirely non-recursive, and the sub-signals are extracted simultaneously; it is proven that VMD outperforms the EMD algorithm in various areas such as signals analysis and damage detection.

Variational mode decomposition has been deployed in the real SHM by some researchers. For instance, Bagheri et al. [31] calculated damping ratios for each extracted modal response obtained from VMD. The mode shape vector was obtained for each decomposed structure mode, which was then practiced for damage identification in three specimens, including numerical, experiment, and field case studies. Xin et al. [33] established two damage indices relying on modal parameters obtained from VMD. An experimental and numerical assessment demonstrated the efficiency of the method for nonlinear to find the location and severity of nonlinear damage scenarios in the models. Das and Saha [34] investigated the impact of a heavy noise environment on a new hybrid algorithm using VMD along with frequency domain decomposition (FDD). It was deducted that the hybrid method could detect damage location accurately for noises above 20%. A novel methodology is illustrated and assessed in the following sections on two experimental specimens with linear and nonlinear damage scenarios.

## IV. Case Studies

In this section, two case studies used in this work are thoroughly explained and discussed.

### *A. Case study 1: Linear Damage*

The first case study is a three-story metal frame with aluminum columns and floors, investigated in linear damage simulation [35]. A roller at the base supports the specimen and can move horizontally using a hydraulic jack. Piezoelectric single-axis accelerometers instrument each floor. Nine linear damage scenarios are imitated employing stiffness reduction of columns and replacement of a 1.2 kg mass. Hence, 50 signals are recorded for each status with a sample rate of 320 Hz. Therefore, 450 signals are acquired for all scenarios, as illustrated in Table 1. As depicted, there are nine statuses, including healthy condition (S1) showing the intact structures without any changes in components, two scenarios simulate the operational and environmental effects by changing mass of floors (S2 and S3), and six damage scenarios by changing the stiffness of columns (S4-S9).

Additionally, Fig. 2 presents sample recorded signals in different scenarios, where $y_1(t)$, $y_2(t)$, and $y_3(t)$ represent the recorded data by sensor 1, sensor 2, and sensor 3, respectively. It is evident that the recorded responses for all damage scenarios follow a random pattern, and usage of time-domain data can not discriminate damage status from healthy cases. Thus, there is a need to model output responses through signal processing techniques find suitable features indicating variations in the signals.

Table 1 Damage scenarios of case study 1 [17]

| Scenario | Records | Description |
|---|---|---|
| S1 | 0-50 | Healthy State |
| S2 | 51-100 | Mass = 1.2 kg at the base |
| S3 | 101-150 | Mass = 1.2 kg on the 1st level |
| S4 | 151-200 | 87.5% stiffness decrease in column 1BD |
| S5 | 201-250 | 87.5% stiffness decrease in column 1AD and 1BD |
| S6 | 251-300 | 87.5% stiffness decrease in column 2BD |
| S7 | 301-350 | 87.5% stiffness decrease in column 2AD and 2BD |
| S8 | 351-400 | 87.5% stiffness decrease in column 3BD |
| S9 | 401-450 | 87.5% stiffness decrease in column 3AD and 3BD |

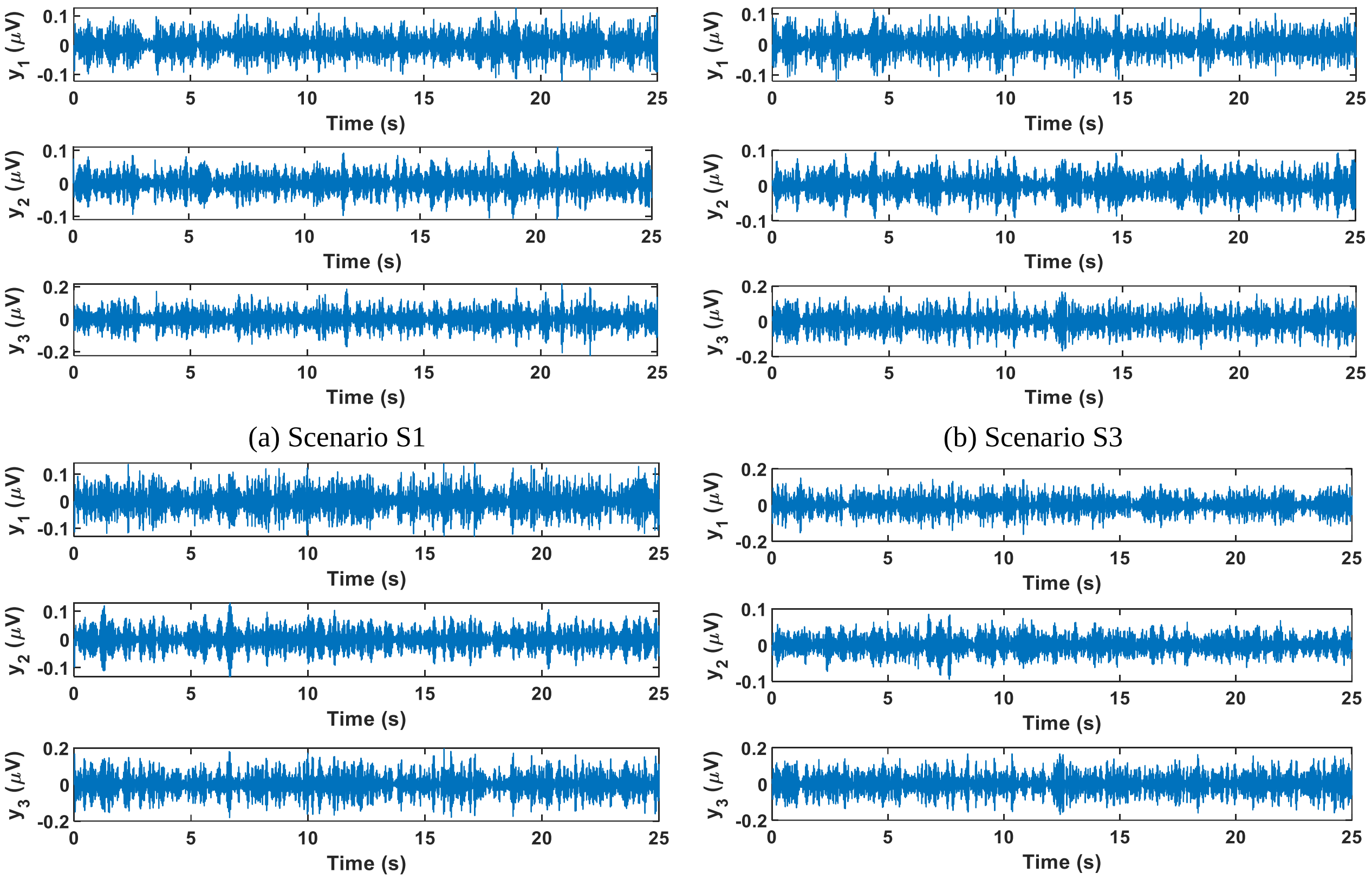

(a) Scenario S1 (b) Scenario S3

(c) Scenario S6 (d) Scenario S9

Fig. 2. Samples from some scenarios in linear damage (scenario 1)

### *B. Case Study 2: Nonlinear Damage*

This case is the adjusted model of the first case study and is used for studying the impact of nonlinear damage. The sampling rate is the same as the linear model and is set to 322.58 Hz with 8192 data points for each record. Ten measurements are recorded for each state. Likewise, in the initial specimen, this frame also glides on rails that enable a transmission in one direction with the aid of an actuator. Four accelerometers with a sensitivity of 1000 mV/g are attached on the opposite side of the shaker at the center of the floors; thus, they do not help determine the specimen's torsion models.

In order to simulate nonlinear damages, a mechanical bumper and a center column are installed onto the frame. This mechanism imitates the breathing crack and will cause nonlinear behaviors in the condition that the installed column hits the bumper, which is placed on the second floor. The adjustable gap between the bumper and the installed column is used for defining different degrees of nonlinearity. Hence, the larger the gap is, the smaller the nonlinear behavior becomes. The specimen's outline and the damage scenarios are provided in Fig. 3 and Table 2, respectively. Some recorded nonlinear signals are given Fig. 4, where $y_1(t)$, $y_2(t)$, $y_3(t)$, and $y_4(t)$ respectively represent the recorded data by sensor 1, sensor 2, sensor 3, and sensor 4. Similar to the previous case, the time-domain presentation of responses can not indicate variations due to damages properly.

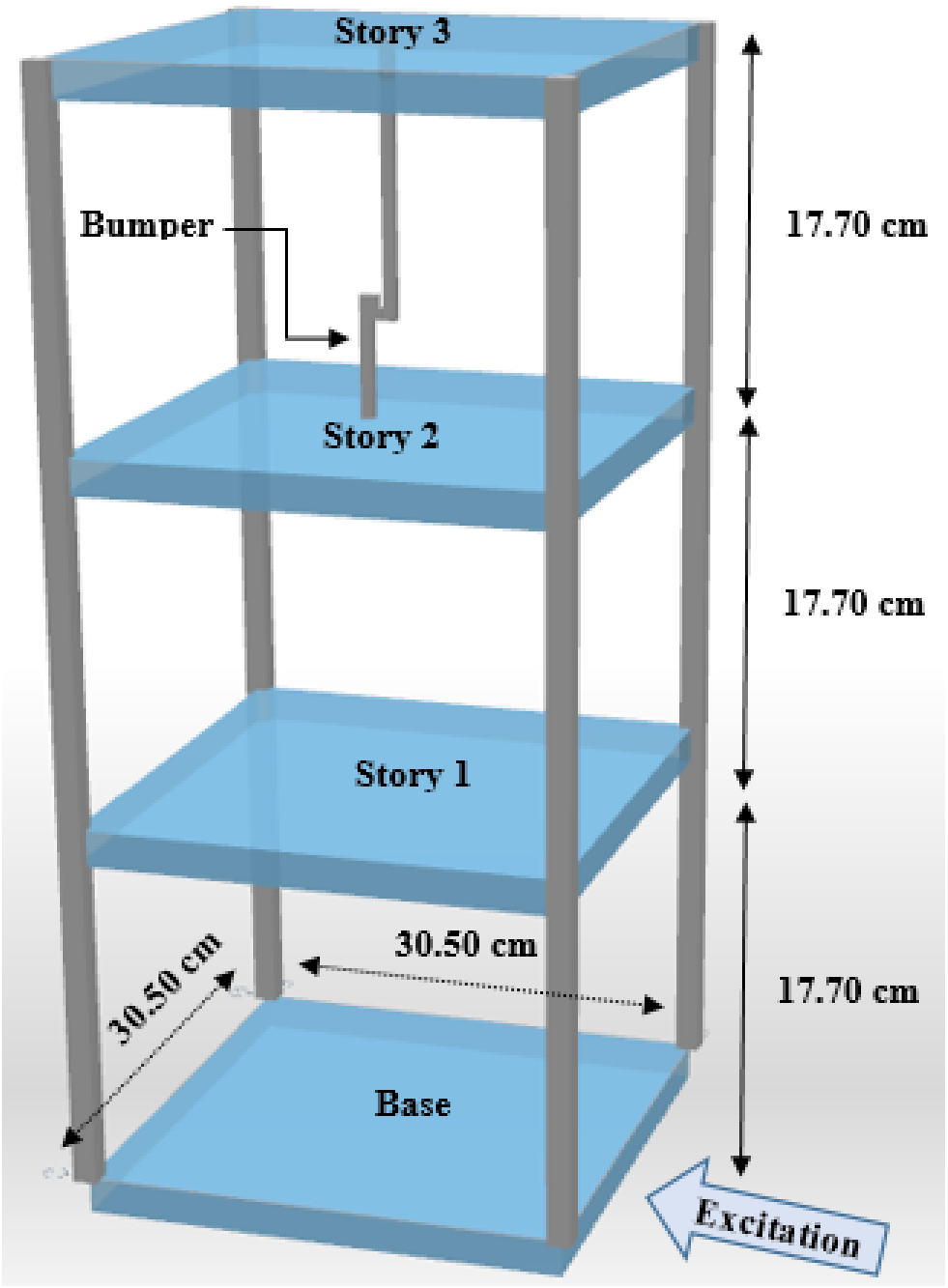

Fig. 3. Three-story bookshelf (adapted from [14]).

Table 2. Damage scenarios of case study 2 [17]

| ***Scenario*** | Records | Description |
|---|---|---|
| **S1** | 1-10 | Mass = 1.2 kg on the 1st floor |
| **S2** | 11-20 | Mass = 1.2 kg at the base |
| **S3** | 21-30 | Gap = 0.13 mm |
| **S4** | 31-40 | Gap = 0.10 mm |
| **S5** | 41-50 | Gap = 0.05 mm |
| **S6** | 51-60 | Gap = 0.15 mm |
| **S7** | 61-70 | Gap = 0.20 mm |
| **S8** | 71-79 | Healthy State |
| **S9** | 80-89 | Gap = 0.20 mm and mass = 1.2 kg at the 1st floor |
| **S10** | 90-99 | Gap = 0.10 mm and mass = 1.2 kg at the 1st floor |
| **S11** | 100-109 | Gap = 0.20 mm and mass = 1.2 kg at the base |
| **S12** | 110-119 | 50.0 % stiffness reduction in column 1BD |
| **S13** | 120-129 | 50.0 % stiffness reduction in column 1AD + 1BD |
| **S14** | 130-139 | 50.0 % stiffness reduction in column 3BD |
| **S15** | 140-149 | 50.0 % stiffness reduction in column 3AD and 3BD |
| **S16** | 150-159 | 50.0 % stiffness reduction in column 2AD and 2BD |
| **S17** | 160-169 | 50.0 % stiffness reduction in column 2BD |

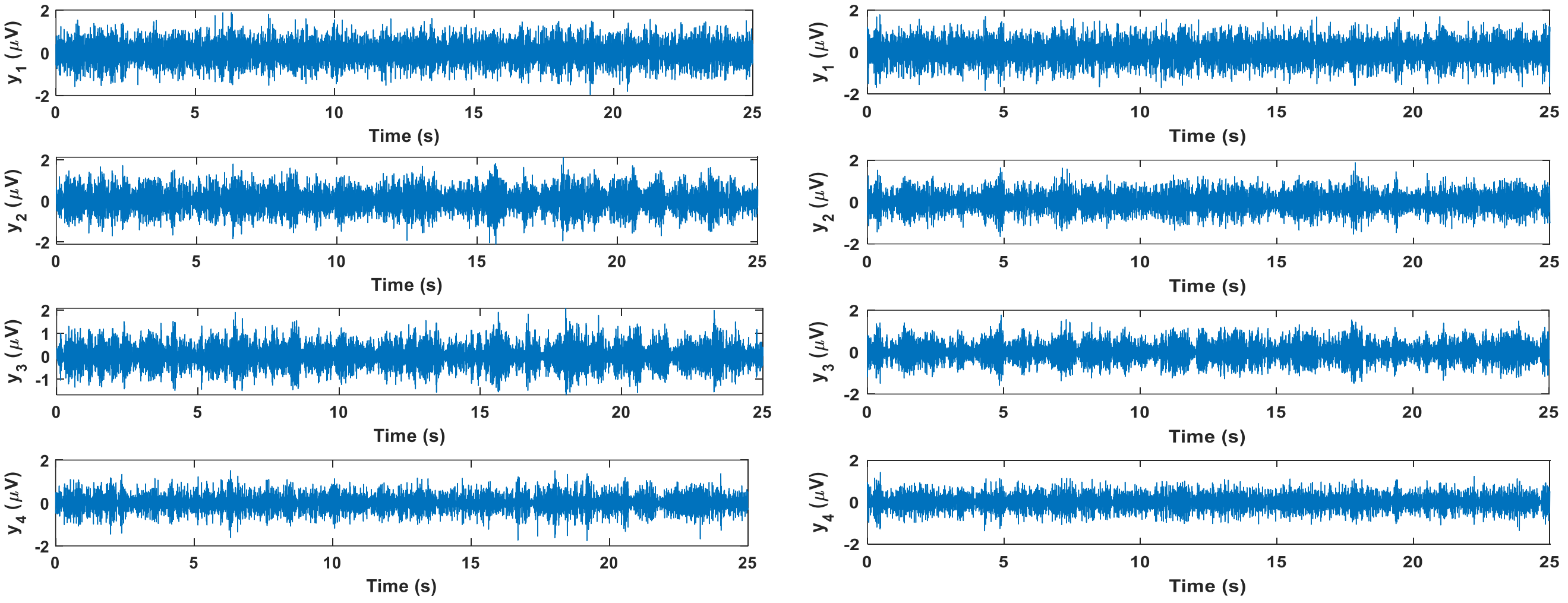

(a) Scenario S7 (b) Scenario S9

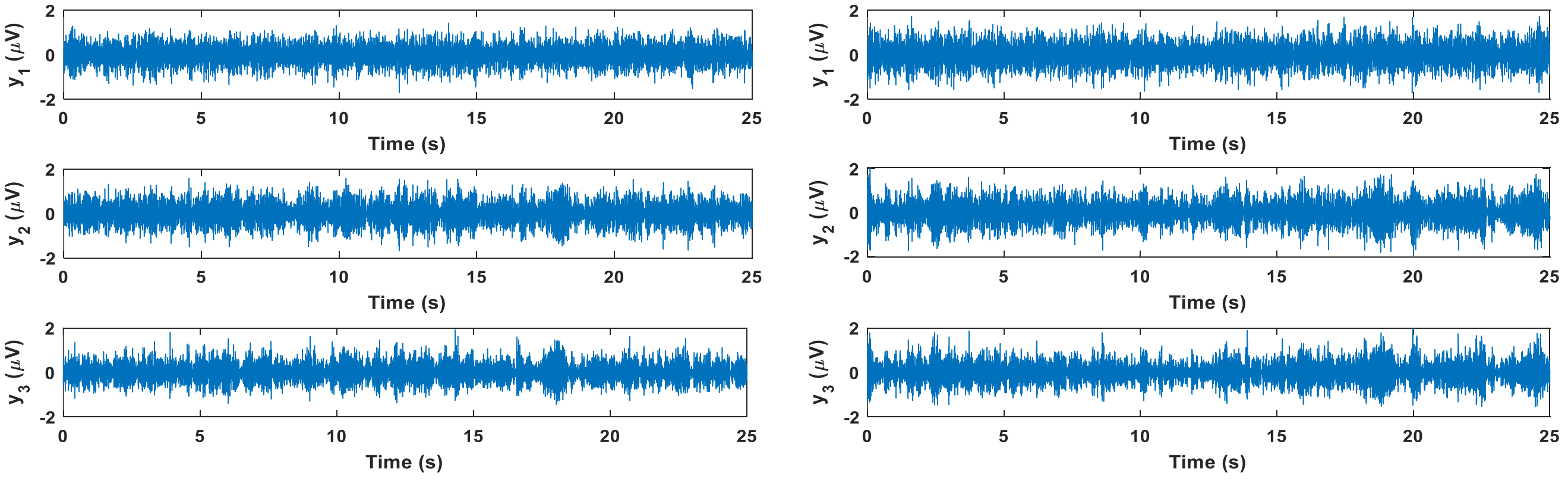

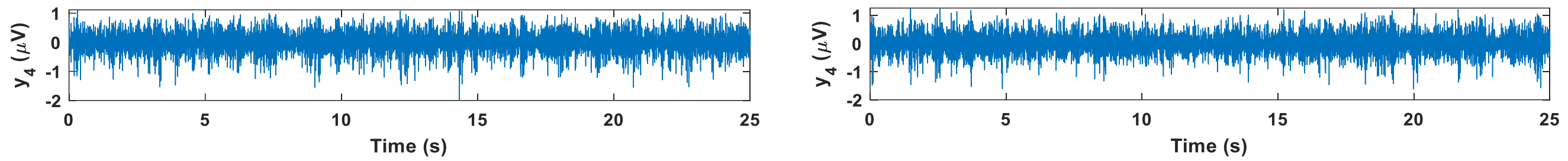

(c) Scenario S11 (d) Scenario S6

Fig. 4. Samples from nonlinear scenarios in case study 2

As noted, two three-story models were presented for linear and nonlinear damage scenarios. Linear damages were simulated by reducing the cross-section area of columns, while nonlinear behavior was considered as hitting a bumper with a mid-column in the second case study. The environmental and operational conditions were also considered by adding a mass to different damage scenarios. Story accelerations were recorded for damage identification and classification, with a novel methodology discussed in the following section.

## V. PROPOSED METHOD

In this work, anomaly detection is performed in three steps. Firstly, VMD decomposes the signal into several sub-signals with separated bandwidths. Secondly, primary features are extracted using the time-series modeling, and then the number of features is reduced by KPCA and KDA. Finally, three supervised classifiers are separately deployed to discriminate different damage states within three specimens. A schematic workflow of the proposed method is depicted in Fig. 5. In the following, these stages are illustrated thoroughly.

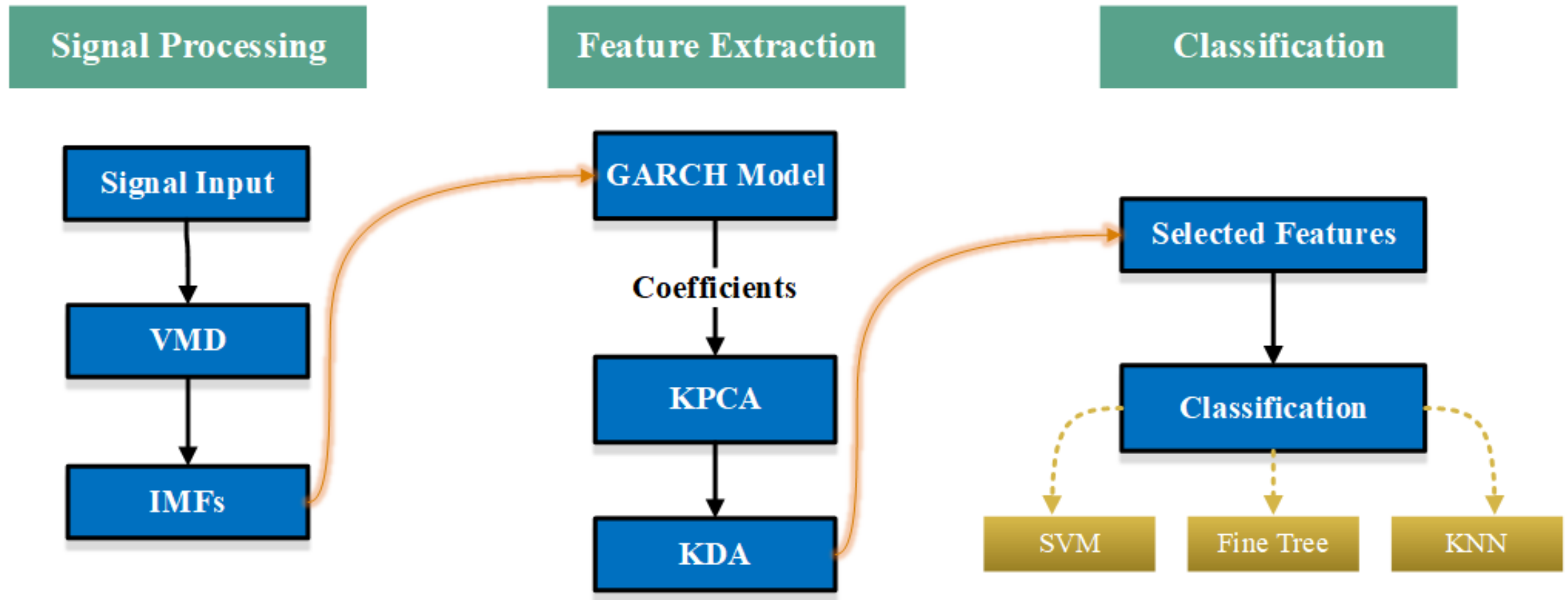

Fig. 5. Workflow of the proposed method

### *A. Signal Processing*

Herein, the input acceleration signals are decomposed using VMD so that an input signal $S(t)$ is broken down into $d$ limited-bandwidth IMFs depicted as [36]:

$$u_k(t) = A_k(t)\cos(\omega_k(t)) \tag{1}$$

where $A_k(t)$ and $\omega_k(t)$ present the instantaneous amplitude and frequency of $u_k(t)$, respectively. The constructed variational problem is obtained using Hilbert transform as follows:

$$\min_{u_k,\omega_k}\left\{\sum_k \left\| \partial_t\left(\sigma(t)+\frac{ju_t(t)}{t}\right)e^{-j\omega_k t}\right\|_2^2\right\} \tag{2}$$

such that

$$S_t = \sum_k u_k(t) \tag{3}$$

where $\partial(t)$ denotes the partial derivative of $t$, $\{u_k(t)\} = \{u_1(t),...,u_n(t)\}$ and $\{\omega_k\} = \{\omega_1,...,\omega_n\}$ shows the IMFs of signal $S_t$ and their center frequencies of each signal sub-band, respectively. Eq. (2) is presented in a Lagrange function using $\lambda$ and $\alpha$ as a multiplier operator and penalty factor, respectively, to solve the optimization problem

$$L\ \{u_k\},\{\omega_k\},\lambda\ = \alpha\sum_k \left\| \partial_t\left(\sigma(t)+\frac{j}{t}u_t(t)\right)e^{-j\omega_k t}\right\|_2^2 + \left\| S(t)-\sum_k u_k(t)\right\|_2^2 + \left\langle \lambda(t), S(t)-\sum_k u_k(t)\right\rangle \tag{4}$$

Afterward, Eq. (4) is transformed into the time-frequency space, and the equivalent extremum solution is solved to obtain the frequency domain form of the modal element $u_k(t)$ as well as the center frequency $\omega_k$:

$$u_k^{n+1}(\omega) = \frac{f(\omega) - \sum_{i=1,i\neq k}^{k} u_i(\omega) + 0.5\lambda(\omega)}{1+2}\alpha(\omega-\omega_k)^2 \tag{5}$$

$$\omega_k^{n+1} = \frac{\int_0^\infty \omega\left|u_k(\omega)\right|^2 d\omega}{\int_0^\infty \left|u_k(\omega)\right|^2 d\omega} \tag{6}$$

Lastly, the alternative direction of multipliers (ADMM) is deployed to optimize the constrained variational model. Subsequently, the initial signal $S(t)$ is broken down by $d$ IMFs as described in the following:

- Initialize the parameters $\{u_k\},\{\omega_k\},\{\lambda^1\}$ and $n \to 0$
- The value of $u_k^{n+1}$ and $\omega_k^{n+1}$ is updated according to (5) and (6).
- The $\lambda^{n+1}$ is updated as stated in:

$$\lambda^n(\omega)+\tau\left(f(\omega)-\sum_k^{n+1} u_k(\omega)\right) \tag{7}$$

- The Eq. (7) is continued till the following criteria are satisfied:

$$\frac{\sum_k \left\| u_k^{n+1}-u_k^n \right\|_2^2}{\left\| u_k^n \right\|_2^2} < \varepsilon \tag{8}$$

Proved that the above condition is met, the iteration procedure stops. Other

Herein, the iteration is stopped; otherwise, it returns to step 2, and $d$ IMFs can be extracted [31, 36]. In Figs. 6-9, the IMFs of linear and nonlinear signals are shown. Due to space limitations, we only present the two IMFs.

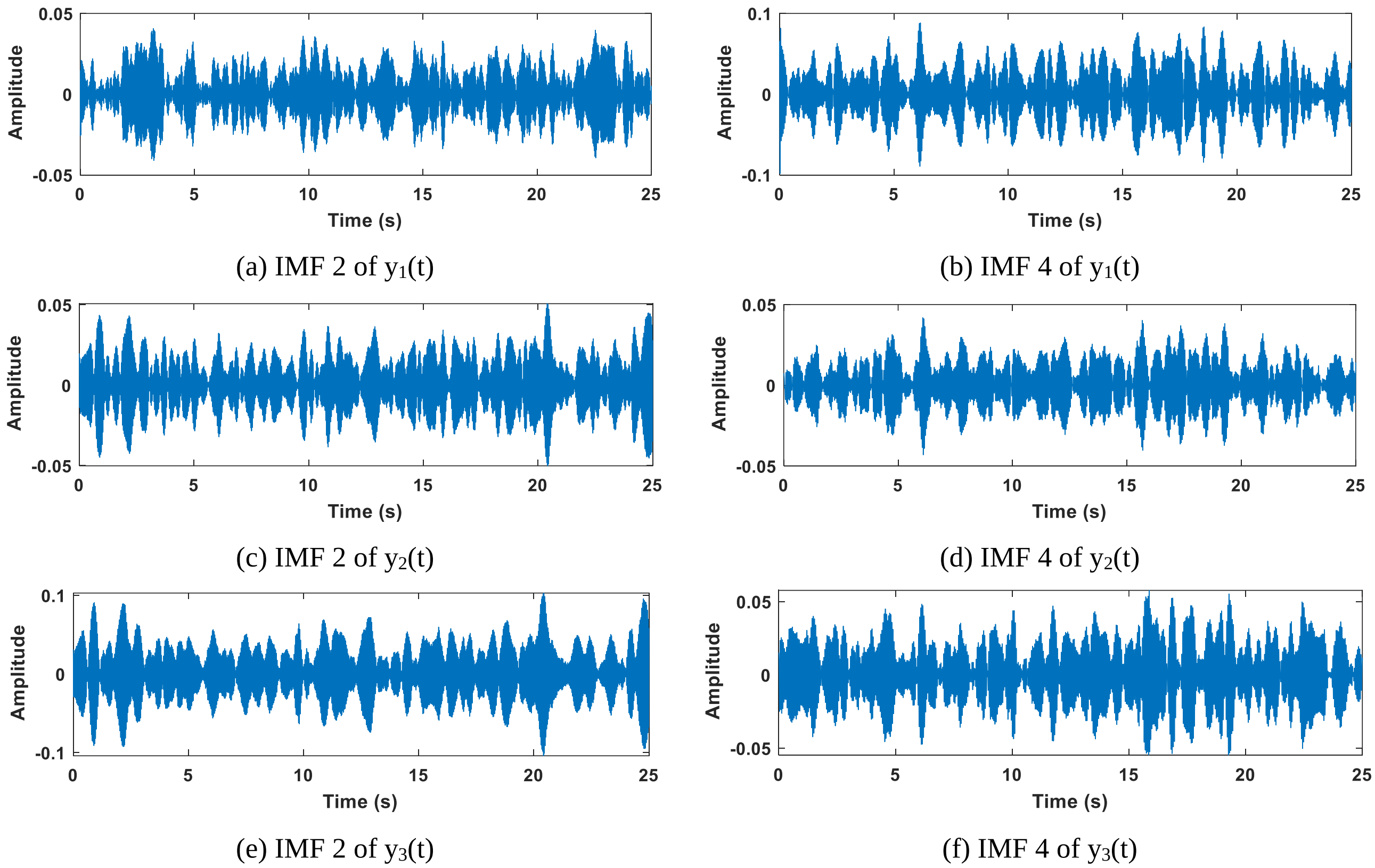

Fig. 6. IMF of signals from scenario S1 (healthy) in case study 1.

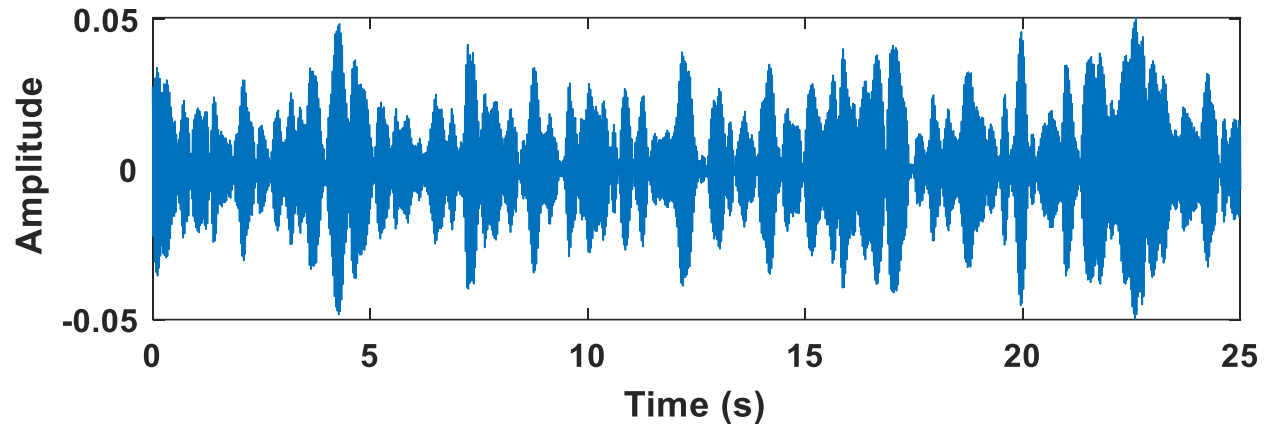

(a) IMF 2 of $y_1(t)$

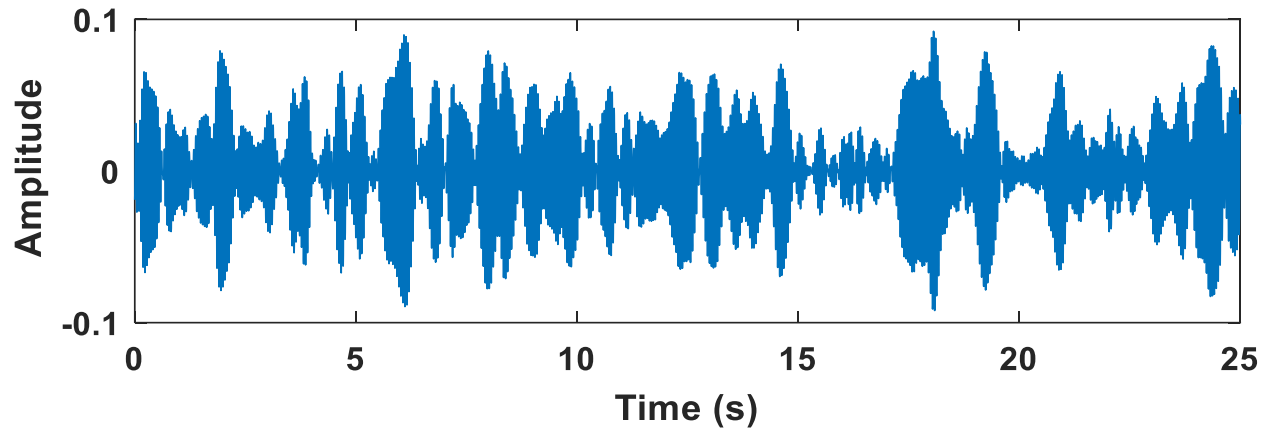

(b) IMF 4 of $y_1(t)$

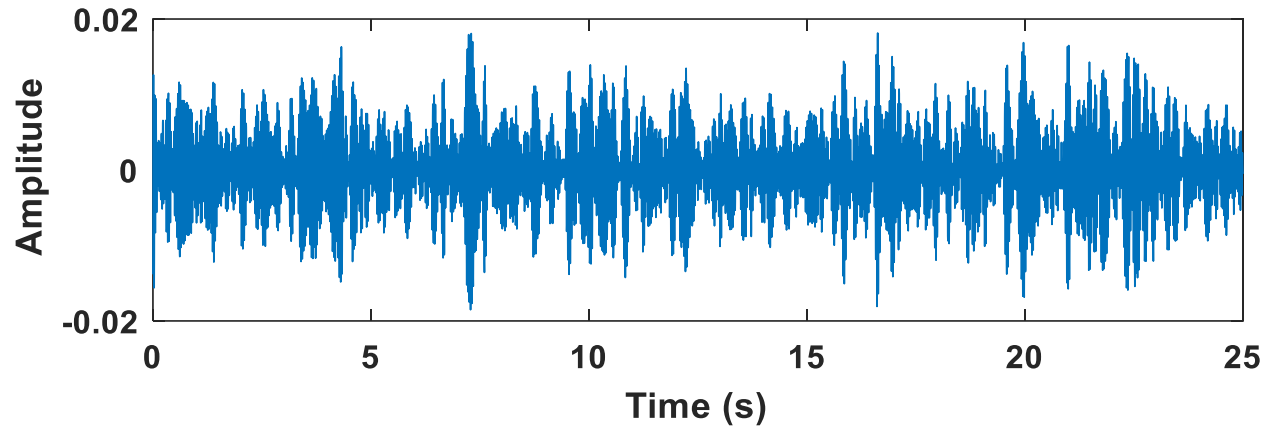

(c) IMF 2 of $y_2(t)$

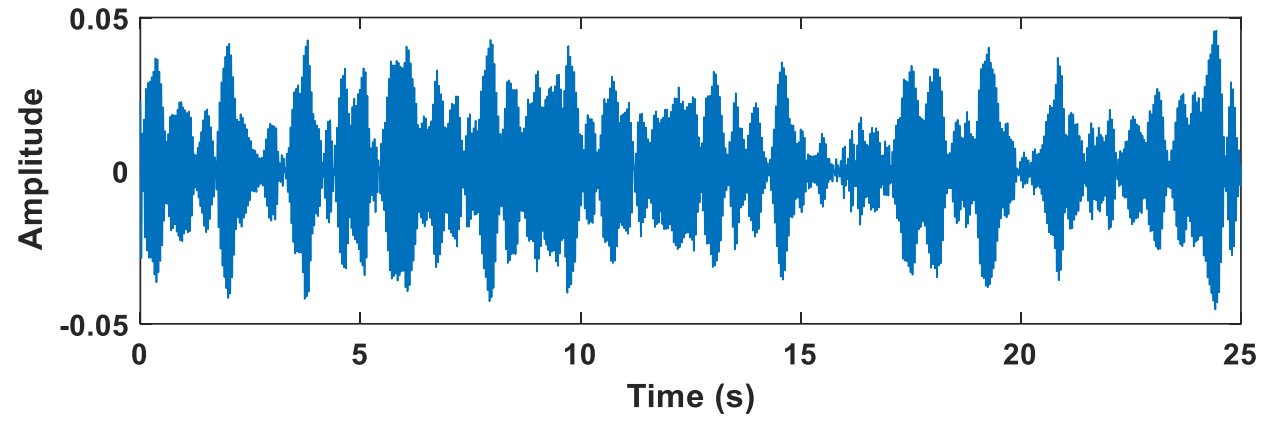

(d) IMF 4 of $y_2(t)$

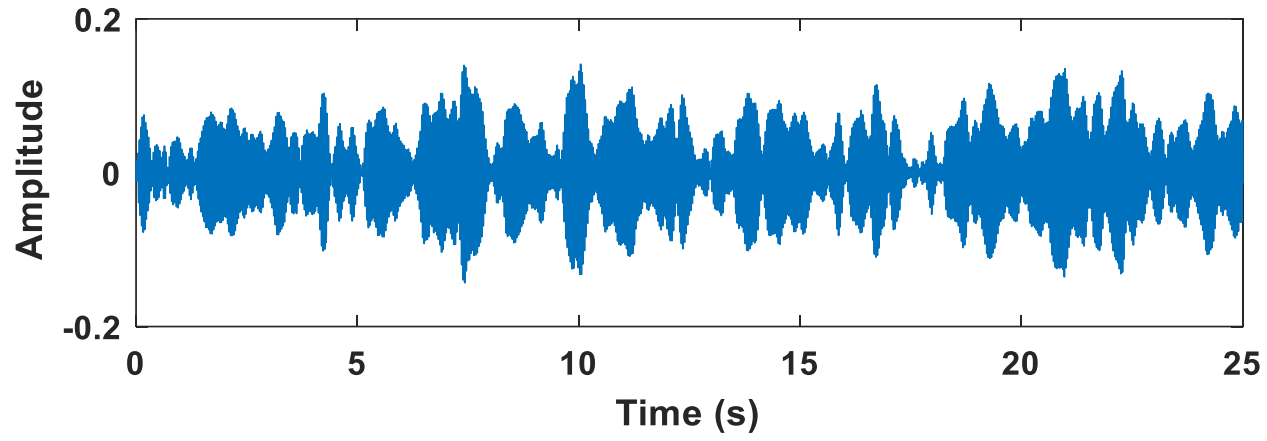

(e) IMF 2 of $y_3(t)$

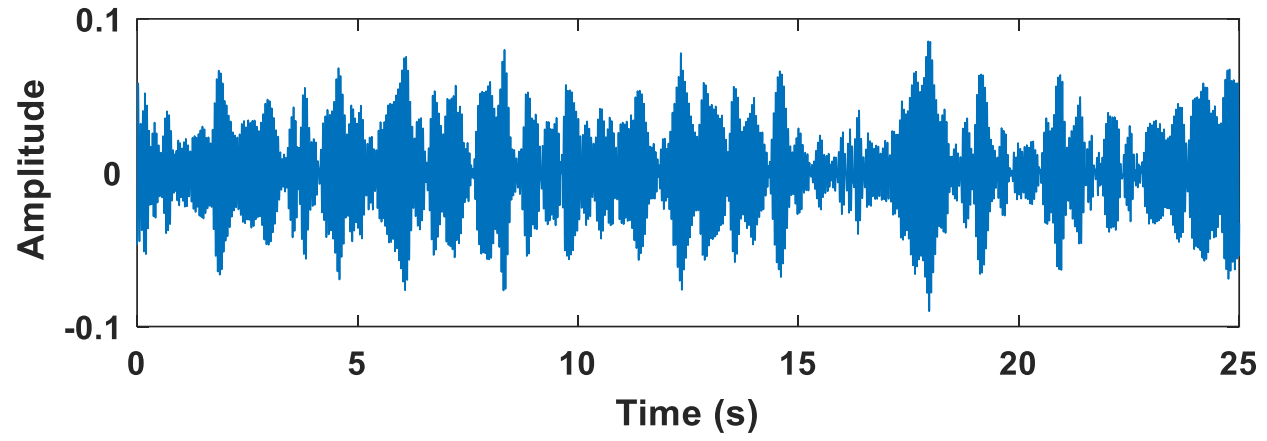

(f) IMF 4 of $y_3(t)$

Fig. 7. IMF of signals from scenario S9 in case study 1.

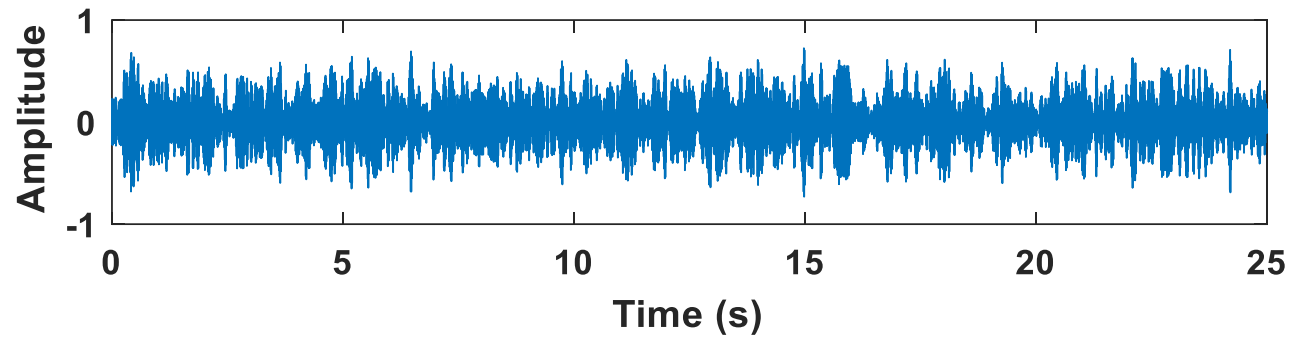

(a) IMF 2 of $y_1(t)$

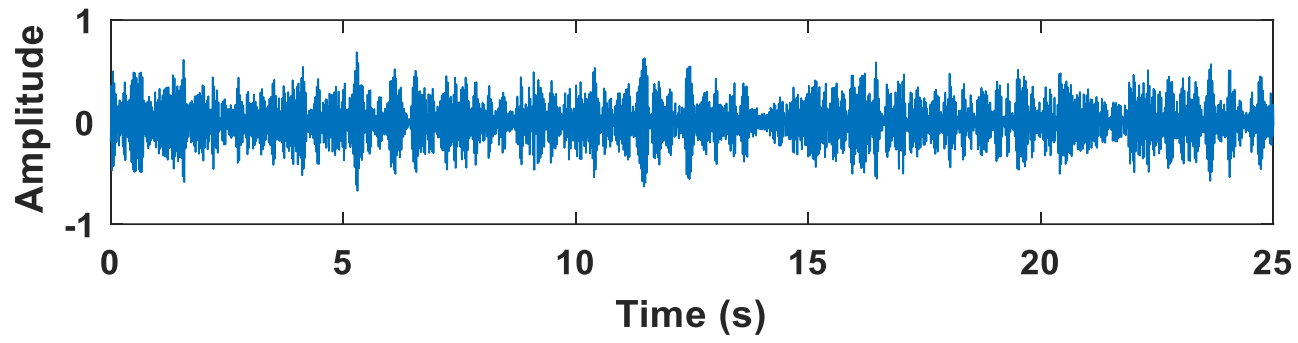

(b) IMF 4 of $y_1(t)$

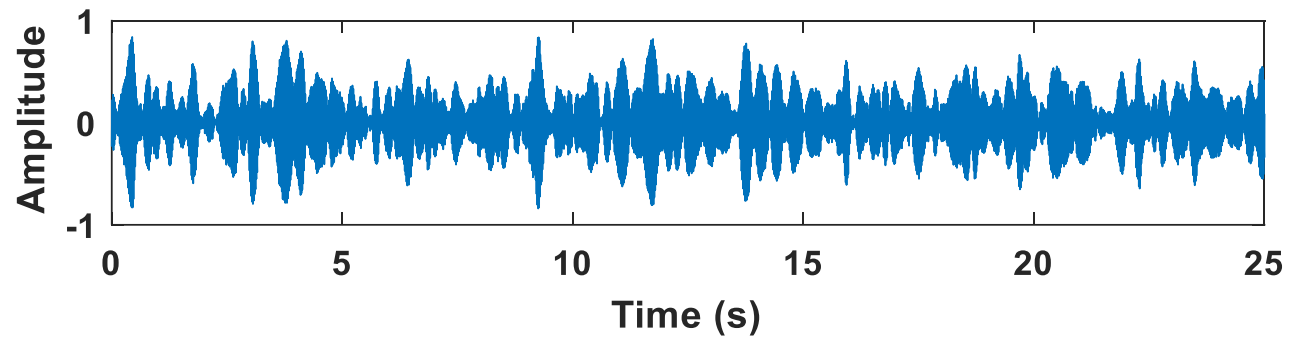

(c) IMF 2 of $y_2(t)$

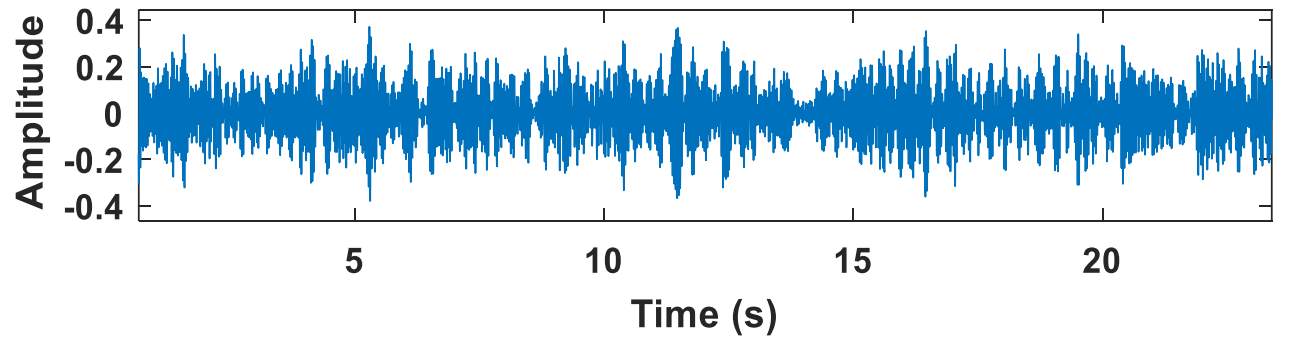

(d) IMF 4 of $y_2(t)$

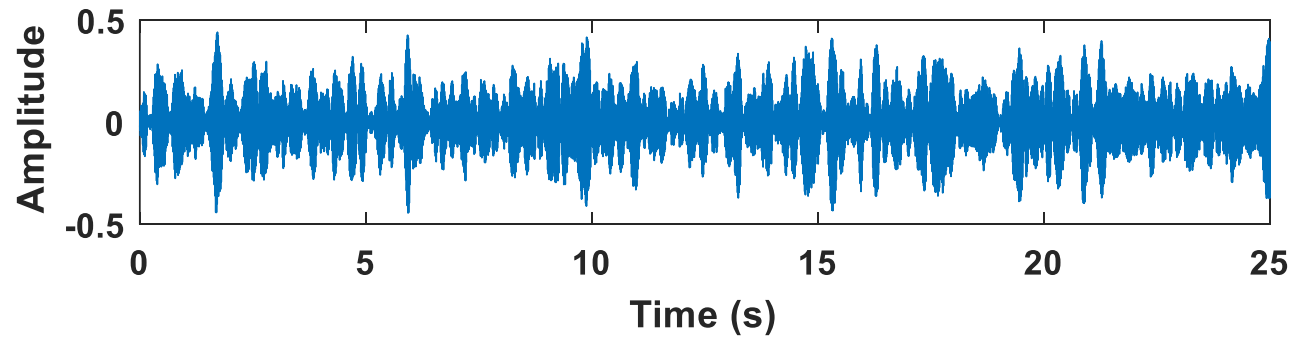

(e) IMF 2 of $y_3(t)$

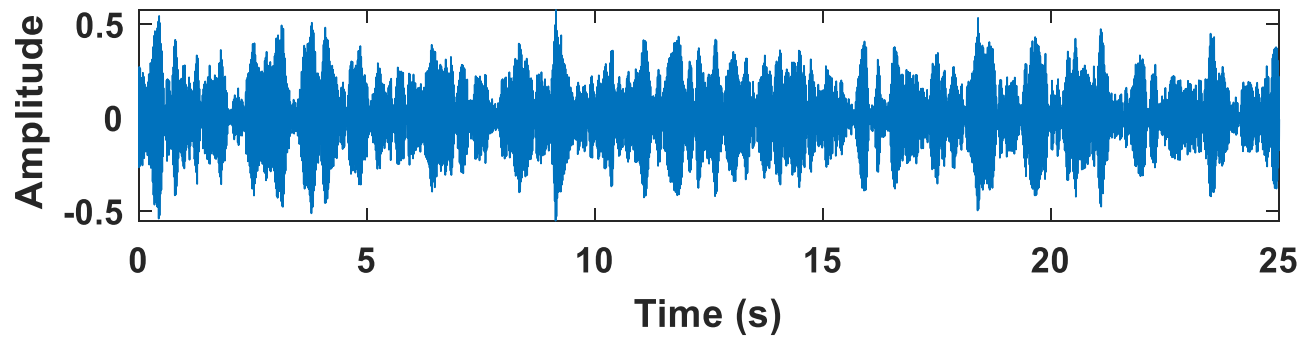

(f) IMF 4 of $y_3(t)$

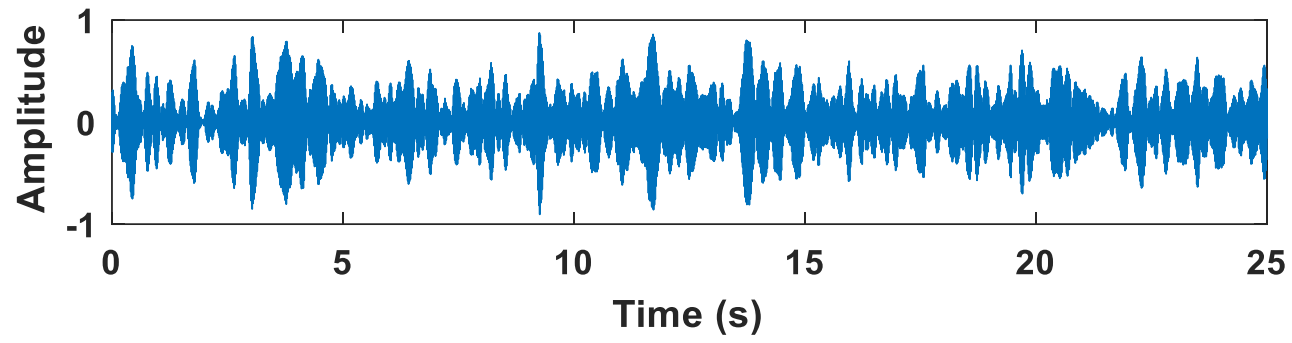

(g) IMF 2 of $y_4(t)$

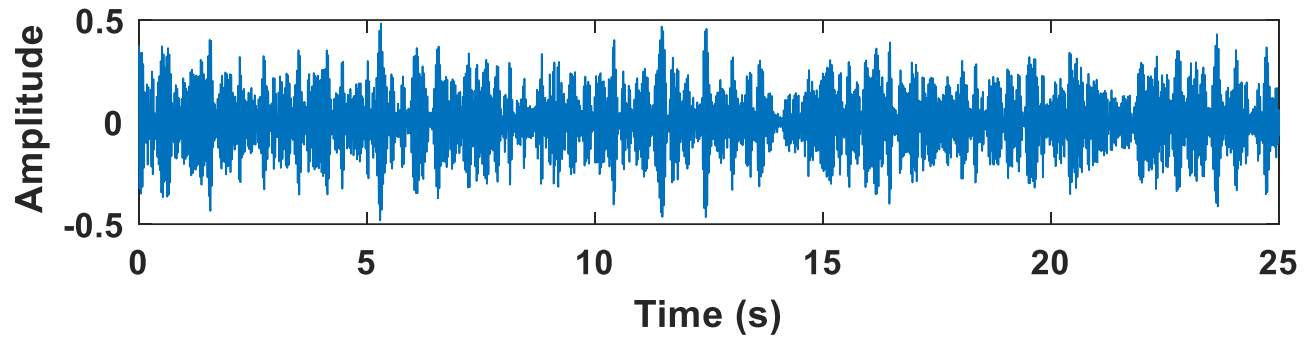

(h) IMF 4 of $y_4(t)$

Fig. 8. IMF of signals from scenario S7 in case study 2.

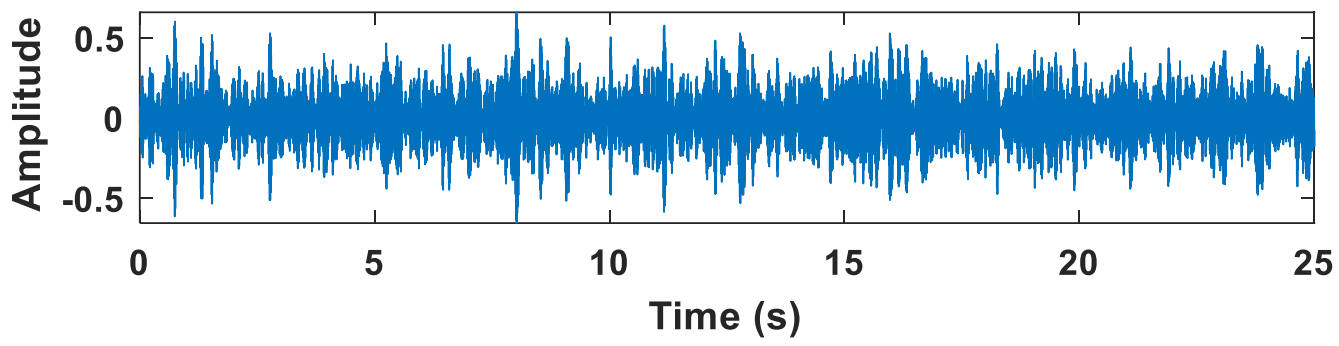

(a) IMF 2 of $y_1(t)$

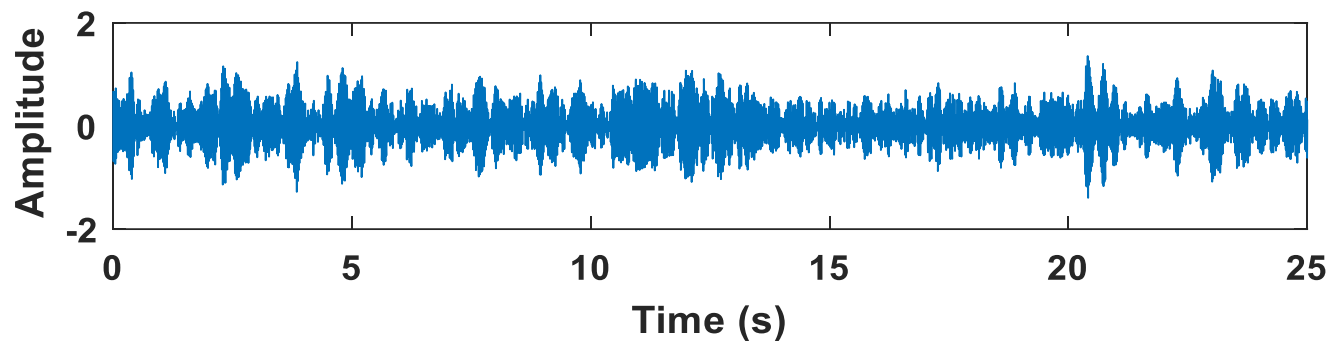

(b) IMF 4 of $y_1(t)$

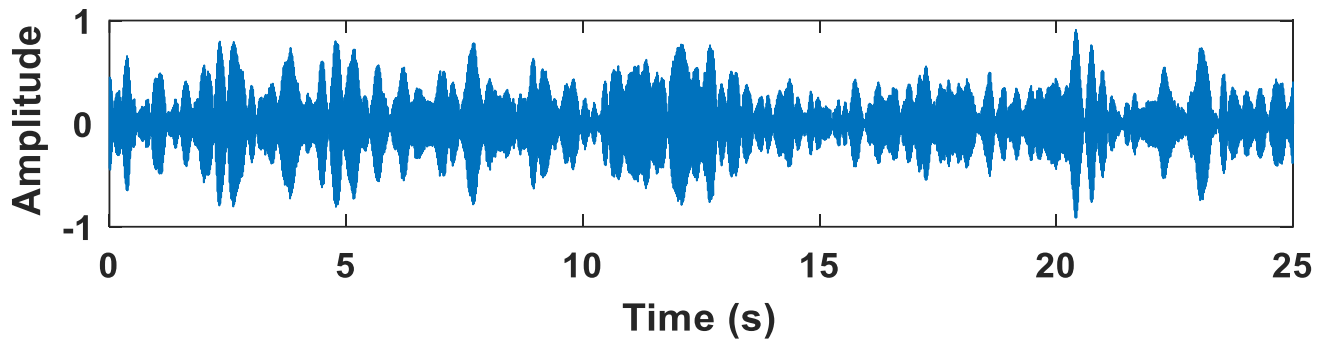

(c) IMF 2 of $y_2(t)$

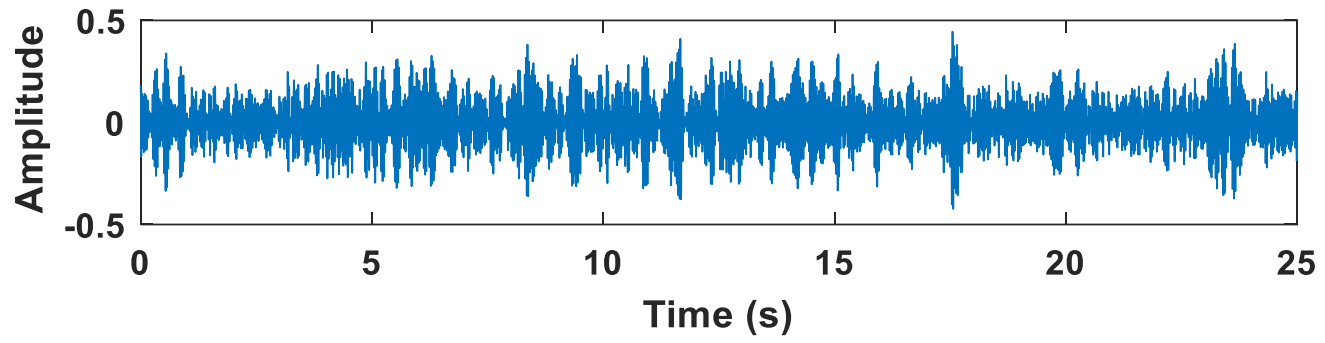

(d) IMF 4 of $y_2(t)$

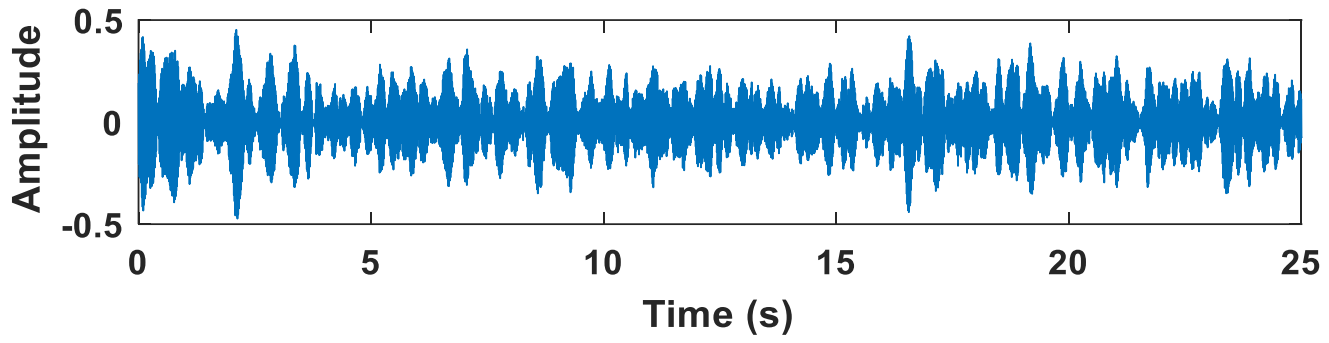

(e) IMF 2 of $y_3(t)$

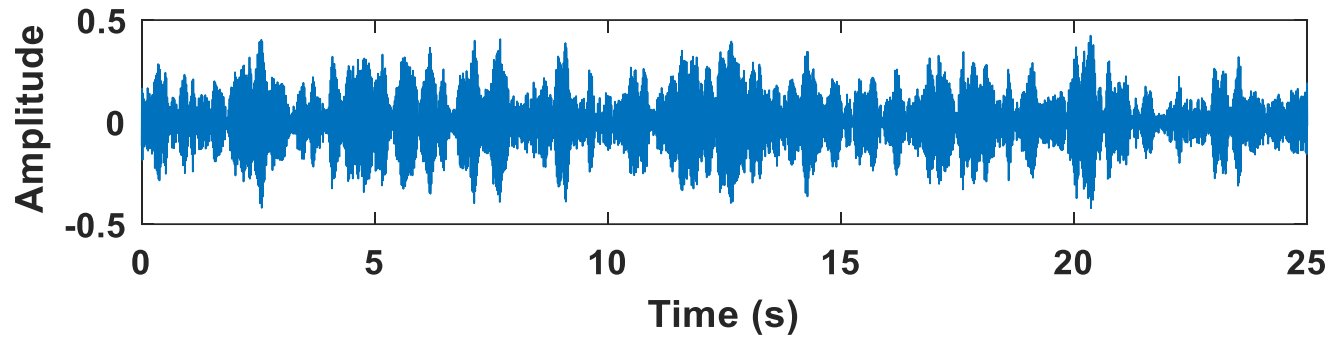

(f) IMF 4 of $y_3(t)$

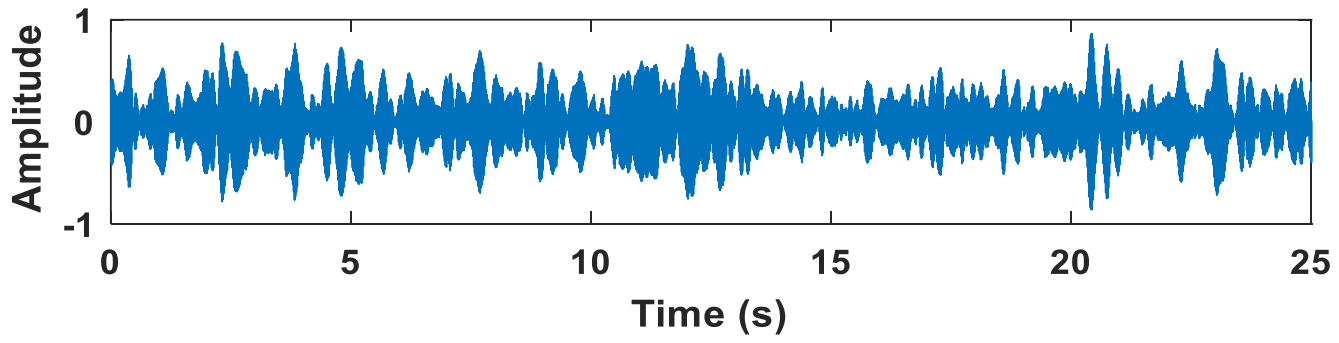

(g) IMF 2 of $y_4(t)$

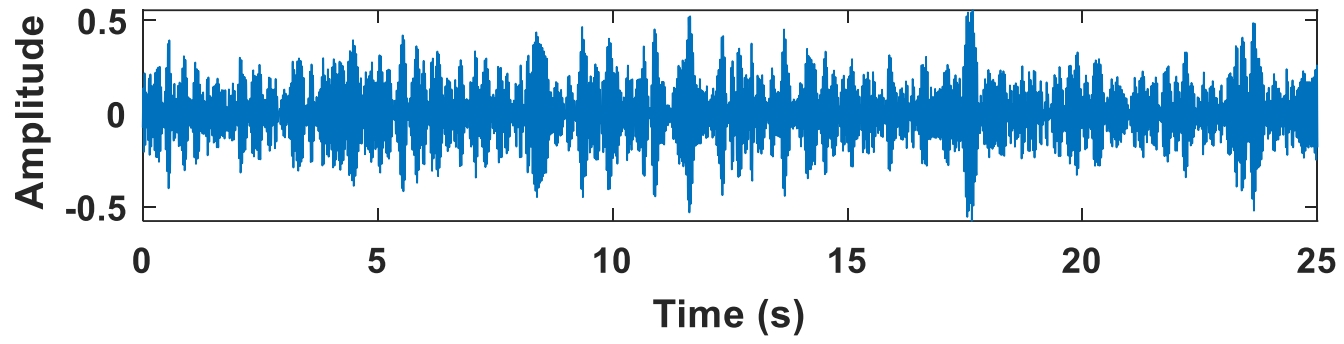

(h) IMF 4 of $y_4(t)$

Fig. 9. IMF of signals from scenario S9 in case study 2.

## *B. Feature Extraction*

### *B.1GARCH modeling of IMFs*

Generally speaking, a signal can be modeled via ARMA time series to evaluate the conditional mean. As an illustration, the ARMA($p$, $q$) prediction for the conditional mean is formulated as [37]:

$$S_t = \sum_{i=1}^{p} \varphi_i S_{t-i} + \sum_{j=1}^{q} \theta_j \varepsilon_{t-j} + \varepsilon_t + c \tag{9}$$

where *p* denotes the autoregressive model order, $\varphi_i$ presents the autoregressive variable, *q* stands for the moving average model order, $\theta_j$ shows the moving average variable, $\varepsilon_t$ denotes the residual, and *c* is a constant. However, the residual is usually considered to have a mean of zero with constant variance. In some time series, it is not homoscedastic and has no constant variance [37]. In this case, the time-varying variance is called conditional variance that is described as:

$$\sigma_t^2 = \mathrm{var}_{t-1}(\varepsilon_t) = E_{t-1} \ \varepsilon_t^2 \tag{10}$$

The GARCH model, established by Bollersl [38], is a dynamic model that addresses the conditional heteroscedasticity or volatility clustering for an innovation process using a weighted combination of past heteroscedasticity functions coupled with the squared residuals of the past. It causes a reduction in the parameters and complexity of the model. A $\mathrm{GARCH}(r,m)$ model for the conditional variance of residual $\varepsilon_t$ is formed as:

$$\sigma_t^2 = \beta + \sum_{i=1}^{r} b_i \sigma_{t-j}^2 + \sum_{j=1}^{m} a_j \varepsilon_{t-j}^2 \tag{11}$$

In which $\beta$, $b_i$, and $a_j$ are the parameters of the GARCH model. Herein, the following constraints are defined to ensure that the conditional variance is positive:

$$\beta > 0, b_i \geq 0, a_j \geq 0 \tag{12}$$

Moreover, the following formula is defined to make the covariance stationary:

$$\sum_{i=1}^{r} b_i + \sum_{j=1}^{m} a_j < 1 \tag{13}$$

This paper utilizes the GRACH model to create the conditional variance model for IMFs obtained from VMD. The GARCH model showed reliable performance in nonlinear problems, as discussed in [19]. The coefficients of $\mathrm{GARCH}(r,m)$, i.e., $\{b_i\}$ and $\{a_j\}$, are considered as features. Hence, *k*th IMF is described by $b_1^{(k)},\ldots,b_r^{(k)},a_1^{(k)},\ldots,a_m^{(k)}$. Considering $d$ IMFs, the feature vector of signal with $d(r+m)$ features, $\mathbf{f}_{(d(r+m))\times 1}$, is constructed as:

$$\mathbf{r} = \left[ b_1^{(1)},\ldots,b_r^{(1)},a_1^{(1)},\ldots,a_m^{(1)},b_1^{(d)},\ldots,b_r^{(d)},a_1^{(d)},\ldots,a_m^{(d)} \right]^{\mathrm{T}} \tag{14}$$

Finally, since each signal is recorded from several sensors, each record is described with $n_f = \sum_{i=1}^{n} d_i(r+m)$ features, where $n$ shows the number of sensors and $d_i$ stands for the number of IMFs is used to decompose the signal of the *i*th sensor. Hence, the feature vector of a signal with $n$ sensors is given as $\mathbf{f} = \left[\mathbf{r}_1^{\mathrm{T}},\ldots,\mathbf{r}_n^{\mathrm{T}}\right]^{\mathrm{T}}$. All obtained features are not suitable for classification, and feature vectors may suffer from redundant features. Hence, we should utilize feature reduction techniques to remove such features from the feature vector.

### *B.2 Feature reduction*

The general concept of kernel-based feature reduction is based on deploying a particular sort of nonlinear mapping function to protrude the initial vector **f** into a high dimensional feature space as *F*. Regarding the new feature space, the principal components are obtained through the regular Principal component analysis (PCA). In other words, the principal nonlinear components in the initial space correspond to the principal components in feature space *F*. Afterward, the kernel functions, including polynomial, radial basis function, and sigmoid, are used to perform the nonlinear mapping in KPCA [39].

Assume nonlinear mapping $\phi$ ; the initial data space $\mathbb{R}^{n_f}$ is mapped into a new feature space like $\mathrm{H}$ as [40]:

$$\begin{aligned} \phi : \mathbb{R}^{n_f} &\rightarrow \mathrm{H} \\ \mathbf{F} &\mapsto \phi(\mathbf{F}) \end{aligned} \tag{15}$$

For a training sample set $\mathbf{f}_1,\mathbf{f}_2,\ldots,\mathbf{f}_M$ in $\mathbb{R}^{n_f}$ , where $M$ denotes training sample numbers. Subsequently, the covariance matrix is formulated as [40]:

$$\mathbf{S}^{\phi} = \frac{1}{M}\sum_{j=1}^{M} \phi(\mathbf{f}_j) - \mathbf{m}_0^{\phi} \quad \phi(\mathbf{f}_j) - \mathbf{m}_0^{\phi} \quad ^{\mathrm{T}} \tag{16}$$

such that

$$\mathbf{m}_0^{\phi} = \frac{1}{M}\sum_{j=1}^{M} \phi(\mathbf{f}_j) \tag{17}$$

Since $\mathbf{S}^{\phi}$ it is a bounded, compact, positive, and symmetric matrix, its nonzero values are also positive. For the sake of finding theses nonzero values, Schölkopf et al. [41] suggested linearly express every eigenvector of $\mathbf{S}^{\phi}$ by [40]:

$$\beta = \sum_{i=1}^{M} \alpha_i \phi(\mathbf{f}_i) \tag{18}$$

In order to compute expansion coefficients, the Gram matrix is formed as $\tilde{R} = \mathbf{Q}^{\mathrm{T}}\mathbf{Q}$ , where $\mathbf{Q} = [\phi(\mathbf{F}_1),...,\phi(\mathbf{F}_M)]$. Consequently, each component $\mathbf{Q}$ is computed by using kernel tricks as [40]:

$$\tilde{R}_{ij} = \phi\left(\mathbf{f}_i\right)^{\mathrm{T}} \phi\left(\mathbf{f}_j\right) = \left(\phi(\mathbf{f}_i).\phi(\mathbf{f}_j)\right) = K\left(\mathbf{f}_i,\mathbf{f}_j\right) \tag{19}$$

Accordingly, $\tilde{\mathbf{R}}$ is centralized by [40]:

$$\mathbf{R} = \tilde{\mathbf{R}} - \mathbf{1}_M\tilde{\mathbf{R}} - \tilde{\mathbf{R}}\mathbf{1}_M + \mathbf{1}_M\tilde{\mathbf{R}}\mathbf{1}_M \tag{20}$$

where

$$\mathbf{1}_M = \left(\frac{1}{M}\right)_{M\times M} \tag{21}$$

Afterward, the orthonormal eigenvectors $\gamma_1,...,\gamma_{n_p}$ of $\mathbf{R}$ are calculated related to $n_p$ the most significant positive eigenvalues such that $\lambda_1 \geq \lambda_2 \geq ... \geq \lambda_{n_p}$ . Consequently, the orthonormal eigenvectors $\beta_1,\beta_2,...,\beta_{n_p}$ of corresponding $\mathbf{S}^{\phi}$ are obtained via [40]:

$$\beta_j = \frac{1}{\sqrt{\lambda_j}} Q\gamma_j \quad ; \quad j = 1,...,n_p \tag{22}$$

After that, the KPCA transformed feature $\mathbf{y} = \left(y_1,...,y_{n_p}\right)^{\mathrm{T}}$ vector is obtained by the projection of the mapped sample $\phi(\mathbf{f})$ onto the eigenvector $\beta_1,\beta_2,...,\beta_{n_p}$ as formulated below [40]:

$$\mathbf{y} = \left(\beta_1,\beta_2,...,\beta_{n_p}\right)^{\mathrm{T}} \phi(\mathbf{f}) \tag{23}$$

The training matrix $\mathbf{F} = \left[\mathbf{f}_1^{\mathrm{T}}; \mathbf{f}_2^{\mathrm{T}}; ...; \mathbf{f}_M^{\mathrm{T}}\right]^{\mathrm{T}}$ with the size of $n_f \times M$ is mapped to the matrix $\mathbf{Y} = \left[\mathbf{y}_1^{\mathrm{T}}; \mathbf{y}_2^{\mathrm{T}}; ...; \mathbf{y}_M^{\mathrm{T}}\right]^{\mathrm{T}}$ with the size of $n_p \times M$.

The aim of linear LDA is as follows[42]:

$$\mathbf{a}_{opt} = \arg\max \frac{\mathbf{a}^{\mathrm{T}}\mathbf{S}_b\mathbf{a}}{\mathbf{a}^{\mathrm{T}}\mathbf{S}_w\mathbf{a}} \tag{24}$$

where $\mathbf{S}_b$ and $\mathbf{S}_w$ reveal the between-class and within-class scatter matrices, which are obtained as:

$$\mathbf{S}_b = \sum_{k=1}^{n_c} m_k \left(\boldsymbol{\mu}^{(k)} - \boldsymbol{\mu}\right)\left(\boldsymbol{\mu}^{(k)} - \boldsymbol{\mu}\right)^{\mathrm{T}} \tag{25}$$

$$\mathbf{S}_w = \sum_{k=1}^{n_c} \left( \sum_{i=1}^{m_k} \left(\mathbf{y}_i^{(k)} - \boldsymbol{\mu}^{(k)}\right)\left(\mathbf{y}_i^{(k)} - \boldsymbol{\mu}^{(k)}\right)^{\mathrm{T}} \right) \tag{26}$$

where $\boldsymbol{\mu}$ is the global mean, $m_k$ stands for the number of samples in the *k*th class, and $\boldsymbol{\mu}^{(k)}$ denotes the mean of the *k*th class. Afterward, the total scatters matrix is defined as $\mathbf{S}_t = \mathbf{S}_b + \mathbf{S}_w$ . The optimum values of **a** correspond to the nonzero eigenvalue of eigenproblem:

$$\mathbf{S}_b\mathbf{a} = \lambda \mathbf{S}_t \mathbf{a} \tag{27}$$

A maximum number of $n_c - 1$ eigenvectors are obtained corresponding to nonzero eigenvalues because the rank of $\mathbf{S}_b$ is limited to $n_c - 1$. Similar mapping (15) is considered to extend the LDA to the nonlinear case. Hence, $\mathbf{S}_b^{\varphi}$ , $\mathbf{S}_w^{\varphi}$, and $\mathbf{S}_t^{\varphi}$ respectively stand for the between-class, within-class, and total scatter matrices in feature space, which are obtained by the following formulation:

$$\mathbf{S}_b^{\varphi} = \sum_{k=1}^{n_c} m_k \left(\boldsymbol{\mu}_{\varphi}^{(k)} - \boldsymbol{\mu}_{\varphi}\right)\left(\boldsymbol{\mu}_{\varphi}^{(k)} - \boldsymbol{\mu}_{\varphi}\right)^{\mathrm{T}} \tag{28}$$

$$\mathbf{S}_w^{\varphi} = \sum_{k=1}^{n_c} \left( \sum_{i=1}^{m_k} \left(\varphi\left(\mathbf{y}_i^{(k)}\right) - \boldsymbol{\mu}_{\varphi}^{(k)}\right)\left(\varphi\left(\mathbf{y}_i^{(k)}\right) - \boldsymbol{\mu}_{\varphi}^{(k)}\right)^{\mathrm{T}} \right) \tag{29}$$

$$\mathbf{S}_t^{\varphi} = \sum_{i=1}^{M} \left(\varphi\left(\mathbf{y}_i\right) - \boldsymbol{\mu}_{\varphi}\right)\left(\varphi\left(\mathbf{y}_i\right) - \boldsymbol{\mu}_{\varphi}\right)^{\mathrm{T}} \tag{30}$$

Assume that $\boldsymbol{\nu}$ shows the projective function in feature space, the associated objective function in feature space is defined as:

$$\boldsymbol{\nu}_{opt} = \arg\max \frac{\boldsymbol{\nu}^T \mathbf{S}_b^{\varphi} \boldsymbol{\nu}}{\boldsymbol{\nu}^T \mathbf{S}_t^{\varphi} \boldsymbol{\nu}} \tag{31}$$

This function can be solved by eigenproblem as:

$$\mathbf{S}_b^{\varphi} \boldsymbol{\nu} = \lambda \mathbf{S}_t^{\varphi} \boldsymbol{\nu} \tag{32}$$

And we have:

$$\boldsymbol{\nu} = \sum_{i=1}^{M} \alpha_i \varphi\left(\mathbf{y}_i\right) \tag{33}$$

Then, we can define an equivalent problem as:

$$\boldsymbol{\alpha}_{opt} = \arg\max \frac{\boldsymbol{\alpha}^{\mathrm{T}} \mathbf{KWK} \boldsymbol{\alpha}}{\boldsymbol{\alpha}^{\mathrm{T}} \boldsymbol{KK} \boldsymbol{\alpha}} \tag{34}$$

where $\boldsymbol{\alpha}=[\alpha_1,...,\alpha_M]^{\mathrm{T}}$. The corresponding eigenproblem is as $\mathbf{KWK}\boldsymbol{\alpha}=\lambda\mathbf{KK}\boldsymbol{\alpha}$, where $\mathbf{K}$ shows the kernel matrix, i.e., $K_{ij}=\kappa(\mathbf{y}_i,\mathbf{y}_j)$ and W is defined as

$$W_{ij}=\begin{cases}1/m_k, & \text{if } \mathbf{y}_i \text{ and } \mathbf{y}_j \text{ both belongs to the } k\text{th class}\\ 0, & \text{otherwise}\end{cases} \tag{35}$$

Each eigenvector $\boldsymbol{\alpha}$ provides a projective function $\boldsymbol{\nu}$ in the feature space. Let $\mathbf{y}$ a data, then we have :

$$\langle \boldsymbol{\nu},\varphi(\boldsymbol{y})\rangle=\sum_{i=1}^{m}\alpha_i\langle\varphi(\boldsymbol{y}_i),\varphi(\boldsymbol{y})\rangle=\sum_{i=1}^{m}\alpha_i\kappa\ \boldsymbol{y}_i,\boldsymbol{y}\ =\boldsymbol{\alpha}^T\kappa\ :,\boldsymbol{y} \tag{36}$$

where $\kappa\ :,\boldsymbol{y}\ \doteq\left[\kappa\ \boldsymbol{y}_1,\boldsymbol{y}\ ,...,\kappa\ \boldsymbol{y}_m,\boldsymbol{y}\ \right]^{\mathrm{T}}$. Let $\boldsymbol{\alpha}_1,...,\boldsymbol{\alpha}_{n_c-1}$ be the $n_c-1$ eigenvectors of the eigenproblem concerning nonzero eigenvalues. The transformation matrix $\Theta=\left[\boldsymbol{\alpha}_1,...,\boldsymbol{\alpha}_{n_c-1}\right]$ is $M\times(n_c-1)$ a matrix that embeds the data sample $\mathbf{y}$ into $n_c-1$ dimensional subspace by

$$\mathbf{y}\rightarrow\mathbf{z}=\Theta^{\mathrm{T}}\kappa(:,\mathbf{y}) \tag{37}$$

***C. Classification***

In the next section, three classifiers are applied to the selected features previously taken and are called predictors. These classifiers are prevailing in the realm of Machine Learning, including support vector machine (SVM), fine tree, and *k*-nearest neighbor (*k*NN). SVM is a supervised training algorithm founded on the fact that measurements can be considered two-dimensional space. Each sample denotes a data point in the space and can be separated by a line in the case of a two-dimensional problem and a plane in the case of the dimensional system [43]. Regarding *k*NN, despite its simplicity, it is common in terms of suing in large training datasets. It allocates an estimated value to a new sample on the ground of plurality or weighted of the k nearest neighbors in the training set [44]. Classification using a decision tree (fine tree) algorithm is very fast and suitable for high-dimensional classification problems. A fine tree is a predictive algorithm mapping from samples about an item to conclusions about its target value. In this model, leaves represent the labels, nodes are the features, and branches denote the junction of features, resulting in label classification [45]. Subsequently, the prediction using these classifiers is compared with each in the following sections.

## VI. RESULTS AND DISCUSSION

This section provides the experimental results and relevant discussions. We considered the five-fold cross-validation to assess the performance of the proposed method. To this end, data were randomly partitioned into five equal-sized groups, and then, the training and test procedure was repeated for five trials. One group was considered for testing data in each trial, and other groups were used to train the classifier. Finally, results were averaged. A. The Effect of the Number of IMFs on Residual

The number of IMFs has a considerable effect on the number of extracted features and the complexity of the proposed method. Here, we determine the efficient number of IMFs based on the mean absolute of residuals, shown in Fig. 10 for different numbers of IMFs of nonlinear signals. It is observed that residual generally reduces as the number of IMFs increases. However, the slope of reduction varies for different sensors. The residuals of sensors 2, 3, and 4 reduce faster than that of sensor 1. As observed, the residual of sensor one does not have a significant variation when the number of IMFs is more than ten. On the other side, the reduction in residuals of sensors 2, 3, and 4 is not notable for the number of IMFs greater than seven. Hence, we consider the ten IMFs for sensor one and seven IMFs for the remaining sensors. Considering 31 IMFs and two features extracted from each IMF, each recording is described with 62 features.

Following the linear case, as observed in Fig. 11, the residuals of all sensors dwindle gradually at nearly the same pace. For any figures over eight IMFs, the residual does not show significant deviations. Thus, for the linear signals, the eight values of IMFs are assigned for all sensors of stories. Considering two features for each IMF, each record is denoted through 48 features.

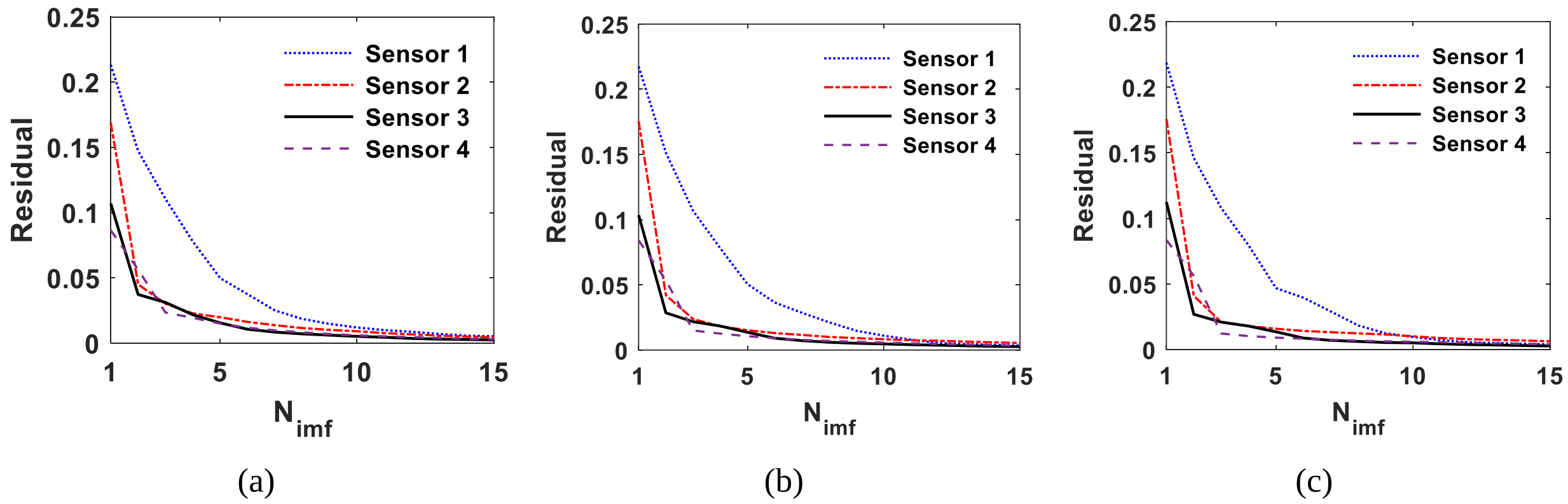

Fig. 10. The residual of VMD of nonlinear data for different numbers of IMFs and data length. (a) length of 512, (b) length of 2048, and (c) length of 8192.

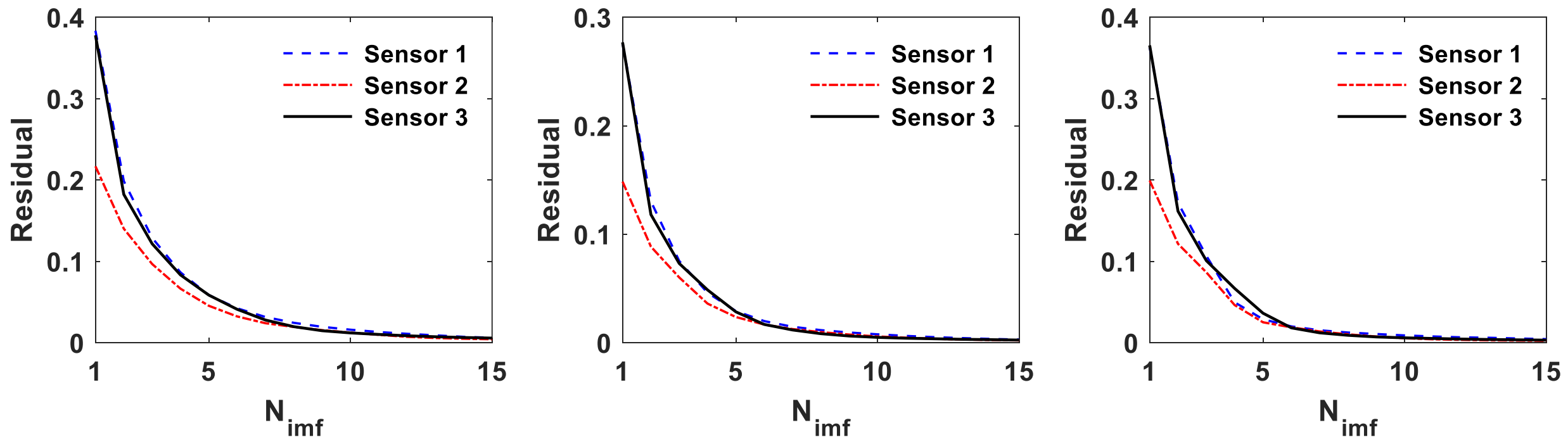

(a) (b) (c)

Fig. 11. The residual of VMD of linear data for different numbers of IMFs and data length. (a) length of 512, (b) length of 2048, and (c) length of 8192.

### *B. Classification Accuracy*

In order to assess the stability of the proposed method and evaluate the effect of features on results, the authors considered four cases as follows:

- SA: no feature reduction method is employed
- SB: only KPCA is used for feature reduction
- SC: only KDA is employed for feature reduction
- SD: at first, KPCA and then KDA is considered for feature reduction.

The number of features in conditions SB, SC, and SD is obtained based on the normalized cumulative summation of eigenvalues (NCSE). When the NCSE reaches higher than 0.95 for the first time, the efficient number of features is obtained. Considering $\left[\lambda_1,\cdots,\lambda_{n_f}\right]$ as sorted eigenvalues in descending order, the NSCE is calculated as follows:

$$\Lambda_i = \frac{\sum_{k=1}^{i}\lambda_k}{\sum_{k=1}^{n_f}\lambda_k} \quad ; \; i = 1,\cdots,n_f$$

Classification accuracy of the proposed method for nonlinear and linear data considering *k*NN, SVM, and fine tree classifiers and different lengths of signals obtained from sensors are given in Table 3 and Table 4, respectively.

Concerning the nonlinear case, the minimum and maximum performance observe in scenario SA and SD with 76.92% and 98.82%, respectively. In all scenarios, fine tree classifiers seem to be more efficient compared to the other classifiers. Moreover, *k*NN is the second accurate classifier, and SVM indicates the lowest performance in this case. It is noteworthy that the signal length has the higher impact on the SB and the lowest on SD with the relative variation ($\Delta_{max}$) of 9.09% and 3.69%, respectively.

Regarding the nonlinear case study, the highest and lowest performance, likewise the nonlinear case, observes in SA and SD with the accuracy of 100.0% and 89.56%, respectively. Similar to the previous case, the fine tree is the suitable classifier in all proposed scenarios. Except for the SB, *k*NN indicates higher performance in comparison with SVM. Scenario SB reveals less sensitivity to the signal length, whereas scenario SA shows the highest sensitivity to the signal variations based on $\Delta_{max}$.

Table 3. Classification accuracy of the proposed method for different classifiers and lengths of nonlinear data

| Condition | Number of features | Classifier | Signal length | | | | | Max | Min | $\Delta_{max}$ (%) |
|---|---|---|---|---|---|---|---|---|---|---|
| | | | 512 | 1024 | 2048 | 4096 | 8192 | | | |
| SA | 62 | *k*NN | 79.88 | 82.25 | 84.02 | 84.61 | 85.79 | 85.79 | 79.88 | 6.89 |
| | | SVM | 76.92 | 78.69 | 79.88 | 81.65 | 84.02 | 84.02 | 76.92 | 8.45 |
| | | Fine Tree | 80.47 | 84.02 | 85.21 | 86.98 | 87.57 | 87.57 | 80.47 | 8.11 |
| SB | Variable for different lengths 51, 51, 50, 48, 43 | *k*NN | 82.84 | 83.43 | 85.80 | 88.75 | 91.12 | 91.12 | 82.84 | 9.09 |
| | | SVM | 81.65 | 84.61 | 85.80 | 87.57 | 88.75 | 88.75 | 81.65 | 8.00 |
| | | Fine Tree | 85.79 | 88.76 | 90.53 | 92.89 | 93.49 | 93.49 | 85.79 | 8.24 |
| SC | 16 | *k*NN | 91.12 | 91.71 | 94.08 | 94.67 | 94.67 | 94.67 | 91.12 | 3.75 |
| | | SVM | 85.79 | 86.39 | 89.35 | 91.12 | 91.17 | 91.17 | 85.79 | 5.90 |
| | | Fine Tree | 91.71 | 92.89 | 94.67 | 95.85 | 96.45 | 96.45 | 91.71 | 4.91 |
| SD | 16 | *k*NN | 92.89 | 94.05 | 95.27 | 96.45 | 96.45 | 96.45 | 92.89 | 3.69 |
| | | SVM | 89.94 | 94.08 | 95.27 | 95.27 | 95.86 | 95.86 | 89.94 | 6.18 |
| | | Fine Tree | 94.08 | 97.63 | 98.22 | 98.82 | 98.82 | 98.82 | 94.08 | 4.80 |

Table 4. Classification accuracy of the proposed method for different classifiers and lengths of linear data

| Scenario | Number of features | Classifier | Signal length | | | | | Max | Min | $\Delta_{max}$ (%) |
|---|---|---|---|---|---|---|---|---|---|---|
| | | | 512 | 1024 | 2048 | 4096 | 8192 | | | |
| SA | 48 | *k*NN | 90.44 | 91.11 | 91.78 | 92.22 | 93.33 | 93.33 | 90.44 | 3.10 |
| | | SVM | 89.56 | 90.22 | 90.89 | 91.56 | 92.66 | 92.66 | 89.56 | 3.35 |
| | | Fine tree | 90.66 | 91.55 | 92.22 | 93.11 | 93.78 | 93.78 | 90.66 | 3.33 |
| SB | Variable for different lengths 40, 39, 35, 32, 30 | *k*NN | 90.66 | 91.11 | 91.33 | 92.22 | 92.66 | 92.66 | 90.66 | 2.16 |
| | | SVM | 90.89 | 91.33 | 92.44 | 93.33 | 93.78 | 93.78 | 90.89 | 3.08 |
| | | Fine tree | 90.89 | 92.44 | 93.33 | 94 | 95.55 | 95.55 | 90.89 | 4.88 |
| SC | 8 | *k*NN | 91.11 | 94.22 | 96.22 | 97.33 | 100 | 100.00 | 91.11 | 8.89 |
| | | SVM | 90.22 | 91.33 | 92.44 | 94.67 | 98.22 | 98.22 | 90.22 | 8.14 |

| | | | | | | | | | |
|---|---|---|---|---|---|---|---|---|---|
| | | Fine tree | 95.77 | 97.77 | 99.33 | 100 | 100 | 100.00 | 95.77 | 4.23 |
| SD | 8 | *k*NN | 96.22 | 97.78 | 99.56 | 100 | 100 | 100.00 | 96.22 | 3.78 |
| | | SVM | 93.33 | 95.11 | 97.11 | 98.67 | 100 | 100.00 | 93.33 | 6.67 |
| | | Fine tree | 97.66 | 98.89 | 100 | 100 | 100 | 100.00 | 97.66 | 2.34 |

### *C. Confusion Matrix*

In this part, the classification performance for both case studies is provided through confusion matrices. Considering the confusion matrix, we provide the recall or sensitivity (Sens.), precision (Prec.), total accuracy (Acc.), and F-score, which are defined as

$$\text{Sens.} = \frac{TP}{TP + FN} \tag{38}$$

$$\text{Prec.} = \frac{TP}{TP + FP} \tag{39}$$

$$\text{Acc.} = 100\frac{TP + TN}{TP + TN + FP + FN} \tag{40}$$

$$F\text{-score} = 2\frac{\text{Prec.} \times \text{Sens.}}{\text{Prec.} + \text{Sens.}} = \frac{TP}{TP + 0.5(FP + FN)} \tag{41}$$

where *TP*, *TN*, *FP*, and *FN* denote the true positive, true negative, false positive, and false negative, respectively.

The results are given in Table 5 for the linear damage and the performance metrics are computed for the nine scenarios described earlier. As indicated, the proposed method determines all damage scenarios with no errors. Consequently, this approach expresses the highest performance for discriminating linear damages based on reference to this study.

Regarding the nonlinear case study, seventeen separate states of the specimen are predicted through the presented technique, and the results are presented by the confusion matrix as depicted in Table 6. As noted, in the majority of the damage states, the prediction accuracy is 100%. Regarding the remaining cases, which are two out of seventeen scenarios, the classification performance is 90.0%. Subsequently, the established strategy revealed considerable performance for recognizing nonlinear and linear damages with significant precision.

Table 5. Confusion matrix for linear case study

| | **Predicted Class** | **Sens.** | **Prec.** |
|---|---|---|---|

| **Linear Case** | | $S_1$ | S2 | S3 | $S_4$ | $S_5$ | $S_6$ | $S_7$ | $S_8$ | $S_9$ | | |
|---|---|---|---|---|---|---|---|---|---|---|---|---|
| **Actual Class** | $S_1$ | 50 | 0 | 0 | 0 | 0 | 0 | 0 | 0 | 0 | 1 | 1 |
| | $S_2$ | 0 | 50 | 0 | 0 | 0 | 0 | 0 | 0 | 0 | 1 | 1 |
| | $S_3$ | 0 | 0 | 50 | 0 | 0 | 0 | 0 | 0 | 0 | 1 | 1 |
| | $S_4$ | 0 | 0 | 0 | 50 | 0 | 0 | 0 | 0 | 0 | 1 | 1 |
| | $S_5$ | 0 | 0 | 0 | 0 | 50 | 0 | 0 | 0 | 0 | 1 | 1 |
| | $S_6$ | 0 | 0 | 0 | 0 | 0 | 50 | 0 | 0 | 0 | 1 | 1 |
| | $S_7$ | 0 | 0 | 0 | 0 | 0 | 0 | 50 | 0 | 0 | 1 | 1 |
| | $S_8$ | 0 | 0 | 0 | 0 | 0 | 0 | 0 | 50 | 0 | 1 | 1 |
| | $S_9$ | 0 | 0 | 0 | 0 | 0 | 0 | 0 | 0 | 50 | 1 | 1 |
| Total accuracy: 100% ; F-score: 1 | | | | | | | | | | | | |

Table 6. Confusion matrix for nonlinear case study

| **Nonlinear case** | | **Predicted label** | | | | | | | | | | | | | | | | | **Sens.** | **Prec.** |
|---|---|---|---|---|---|---|---|---|---|---|---|---|---|---|---|---|---|---|---|---|
| | | $S_1$ | $S_2$ | $S_3$ | $S_4$ | $S_5$ | $S_6$ | $S_7$ | $S_8$ | $S_9$ | $S_{10}$ | $S_{11}$ | $S_{12}$ | $S_{13}$ | $S_{14}$ | $S_{15}$ | $S_{16}$ | $S_{17}$ | | |
| **Actual label** | $S_1$ | 10 | 0 | 0 | 0 | 0 | 0 | 0 | 0 | 0 | 0 | 0 | 0 | 0 | 0 | 0 | 0 | 0 | 1 | 1 |
| | $S_2$ | 0 | 10 | 0 | 0 | 0 | 0 | 0 | 0 | 0 | 0 | 0 | 0 | 0 | 0 | 0 | 0 | 0 | 1 | 1 |
| | $S_3$ | 0 | 0 | 10 | 0 | 0 | 0 | 0 | 0 | 0 | 0 | 0 | 0 | 0 | 0 | 0 | 0 | 0 | 1 | 1 |
| | $S_4$ | 0 | 0 | 0 | 10 | 0 | 0 | 0 | 0 | 0 | 0 | 0 | 0 | 0 | 0 | 0 | 0 | 0 | 1 | 1 |
| | $S_5$ | 0 | 0 | 0 | 0 | 10 | 0 | 0 | 0 | 0 | 0 | 0 | 0 | 0 | 0 | 0 | 0 | 0 | 1 | 1 |
| | $S_6$ | 0 | 0 | 0 | 0 | 0 | 10 | 0 | 0 | 0 | 0 | 0 | 0 | 0 | 0 | 0 | 0 | 0 | 1 | 1 |
| | $S_7$ | 0 | 0 | 0 | 0 | 0 | 0 | 10 | 0 | 0 | 0 | 0 | 0 | 0 | 0 | 0 | 0 | 0 | 1 | 1 |
| | $S_8$ | 0 | 0 | 0 | 0 | 0 | 0 | 0 | 9 | 0 | 0 | 0 | 0 | 0 | 0 | 0 | 0 | 0 | 1 | 1 |
| | $S_9$ | 0 | 0 | 0 | 0 | 0 | 0 | 0 | 0 | 9 | 0 | 1 | 0 | 0 | 0 | 0 | 0 | 0 | 0.9 | 1 |
| | $S_{10}$ | 0 | 0 | 0 | 0 | 0 | 0 | 0 | 0 | 0 | 10 | 0 | 0 | 0 | 0 | 0 | 0 | 0 | 1 | 0.909 |
| | $S_{11}$ | 0 | 0 | 0 | 0 | 0 | 0 | 0 | 0 | 0 | 0 | 10 | 0 | 0 | 0 | 0 | 0 | 0 | 1 | 0.909 |
| | $S_{12}$ | 0 | 0 | 0 | 0 | 0 | 0 | 0 | 0 | 0 | 1 | 0 | 9 | 0 | 0 | 0 | 0 | 0 | 0.9 | 1 |
| | $S_{13}$ | 0 | 0 | 0 | 0 | 0 | 0 | 0 | 0 | 0 | 0 | 0 | 0 | 10 | 0 | 0 | 0 | 0 | 1 | 1 |
| | $S_{14}$ | 0 | 0 | 0 | 0 | 0 | 0 | 0 | 0 | 0 | 0 | 0 | 0 | 0 | 10 | 0 | 0 | 0 | 1 | 1 |
| | $S_{15}$ | 0 | 0 | 0 | 0 | 0 | 0 | 0 | 0 | 0 | 0 | 0 | 0 | 0 | 0 | 10 | 0 | 0 | 1 | 1 |
| | $S_{16}$ | 0 | 0 | 0 | 0 | 0 | 0 | 0 | 0 | 0 | 0 | 0 | 0 | 0 | 0 | 0 | 10 | 0 | 1 | 1 |
| | $S_{17}$ | 0 | 0 | 0 | 0 | 0 | 0 | 0 | 0 | 0 | 0 | 0 | 0 | 0 | 0 | 0 | 0 | 10 | 1 | 1 |
| Total accuracy: 98.82% ; F-score: 0.988 | | | | | | | | | | | | | | | | | | | | |

### *D. The effect of noise*

The various intensity of noises is applied to the responses on the grounds of signal-noise ratio (SNR) to assess the stability of the proposed method against noise, as depicted in Fig. 12. As observed, the proposed method is efficient even in environments contaminated with severe noise (SNR = 1). Furthermore, the established approach can maintain its performance against noise where it shows insignificant variations in the case of SNR of 20 and 15.

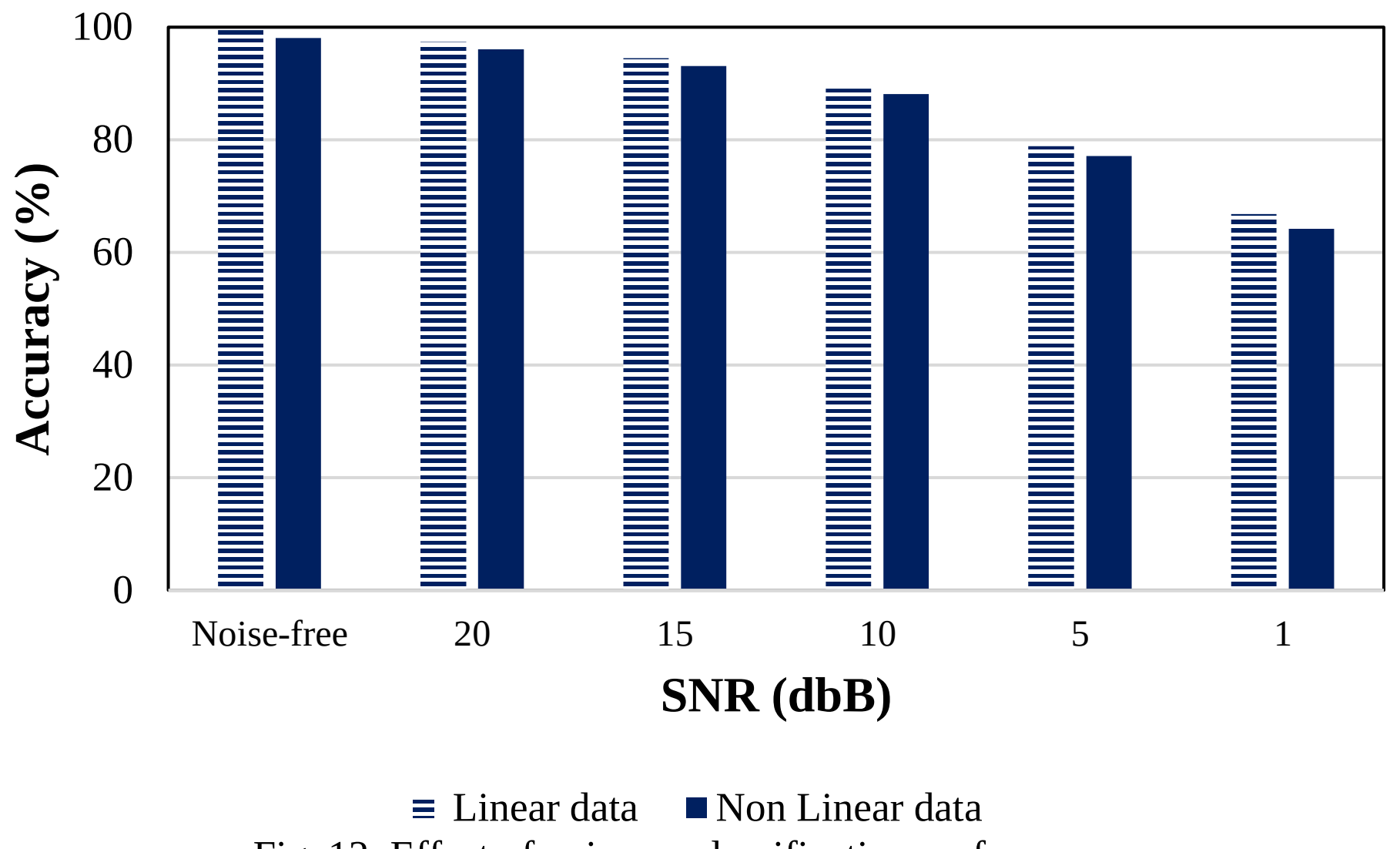

Fig. 12. Effect of noise on classification performance

## VII. GARCH Effect Assessment

In this section, two tests are applied to demonstrate the compatibility of the GARCH model [46]. Thus, Kurtosis and ARCH tests are provided in the following sections.

### *A. Kurtosis test*

GARCH model is appropriate for those signals that have the shape of heavy tails. Therefore, the Kurtosis test is utilized to find out that signals have heavy tails or not. The Kurtosis for a distribution (s) is formulated as follows [46]:

$$K(s) = \frac{E(s-\mu)^4}{\sigma^4} \tag{42}$$

Where $\mu$ and $\sigma$ denote the mean and standard deviation of distribution s, respectively, and $E(s)$ stands for the expected value of s. For Gaussian distribution, the Kurtosis value of three and higher values shows that the distribution of coefficients has a heavier tail than the Gaussian distribution. This paper applies this test to the IMFs for each sensor, and the average results for minimum and maximum values of sub-bands are presented in Table 7. Regarding the results, it can be seen that the maximum values are higher than 3, which proves that the IMFs do not have Gaussian distribution.

Table 7 Kurtosis test for IMFs

| Sensor | Statistic | IMF number | | | | | | | | | |
|---|---|---|---|---|---|---|---|---|---|---|---|
| | | 1 | 2 | 3 | 4 | 5 | 6 | 7 | 8 | 9 | 10 |
| 1 | Min | 2.81 | 2.75 | 2.75 | 2.80 | 2.87 | 2.87 | 2.99 | 2.78 | 2.72 | 2.88 |
| | Max | 3.36 | 3.28 | 3.41 | 3.89 | 3.54 | 3.87 | 4.27 | 4.07 | 3.47 | 4.74 |
| 2 | Min | 2.88 | 2.78 | 2.92 | 2.57 | 2.67 | 2.85 | 2.54 | 2.73 | 2.89 | 2.75 |
| | Max | 4.61 | 6.93 | 8.18 | 4.92 | 6.50 | 4.10 | 3.63 | 3.91 | 3.99 | 6.70 |
| 3 | Min | 2.77 | 2.84 | 2.82 | 2.88 | 2.92 | 2.94 | 2.89 | 2.86 | 2.77 | 2.94 |
| | Max | 24.57 | 23.77 | 8.97 | 4.76 | 4.62 | 4.38 | 3.91 | 6.02 | 8.85 | 4.83 |
| **4** | Min | 2.88 | 2.98 | 2.91 | 2.92 | 2.81 | 2.93 | 2.93 | 2.93 | 2.87 | 2.97 |
| | Max | 41.64 | 8.66 | 4.78 | 4.68 | 4.23 | 3.93 | 3.61 | 4.08 | 4.277 | 4.24 |

### *B. ARCH test*

Based on the hypothesis provided in [47], the ARCH test is deployed to see the existence of ARCH/GARCH impact in the IMFs of each sensor. In this reference, the Lagrange multiplier test is presented based on regression. Subsequently, the test statistic is asymptotically Chi-square distributed has q degrees of freedom[46].

Thus, in this part, the ARCH test is applied to the IMFs for different sub-bands, and the average results for signals are shown in Table 8. In this table, h stands for the Boolean decision variable, where 1 shows the rejection of the null hypothesis, which depicts that no GARCH effect exists. pvalue is the significance level at which the test rejects the null hypothesis. GARCHstat and CriticalValue are the ARCH test static and critical values of the Chi-square distribution, respectively. Based on this test, if GARCHstat is less than the critical value, no GARCH effect exists. In this study, the significance level is set to 0.05, frequently deployed in [48]. Notably, these results are the average of all signals; for example, the average value of h for the fourth IMF of the first sensor is 0.74, which demonstrates that 74% of the signals have the GARCH effect. Thus, in general, the results of the table prove the existence of the GARCH effect in most cases.

Table 8 ARCH test for IMFs

| Sensor | Results of ARCH test | IMF number | | | | | | | | | |
|---|---|---|---|---|---|---|---|---|---|---|---|
| | | 1 | 2 | 3 | 4 | 5 | 6 | 7 | 8 | 9 | 10 |
| **1** | *h* | 1 | 1 | 0.95 | 0.74 | 0.66 | 0.76 | 0.94 | 1 | 1 | 1 |
| | *pValue* | 0 | 0 | 0.01 | 0.09 | 0.17 | 0.11 | 0.01 | 0 | 0 | 0 |
| | *GARCHstat* | 4416.8 | 1532.5 | 300.2 | 33.8 | 26.2 | 95.1 | 276.8 | 695.1 | 1667.1 | 3923.4 |
| | *CriticalValue* | 3.84 | 3.84 | 3.84 | 3.84 | 3.84 | 3.84 | 3.84 | 3.84 | 3.84 | 3.84 |
| **2** | *h* | 0.62 | 0.65 | 0.92 | 0.97 | 1 | 1 | 1 | 1 | 1 | 1 |
| | *pValue* | 0.19 | 0.14 | 0.03 | 0.01 | 0 | 0 | 0 | 0 | 0 | 0 |
| | *GARCHstat* | 62.19 | 77.38 | 366.2 | 503.6 | 761.6 | 1023.2 | 1503.2 | 2280.3 | 3219.2 | 4629.9 |
| | *CriticalValue* | 3.84 | 3.84 | 3.84 | 3.84 | 3.84 | 3.84 | 3.84 | 3.84 | 3.84 | 3.84 |
| **3** | *h* | 0.83 | 0.78 | 0.93 | 0.98 | 0.95 | 0.97 | 0.99 | 1 | 1 | 1 |
| | *pValue* | 0.07 | 0.08 | 0.02 | 0.01 | 0.01 | 0.01 | 0.01 | 0 | 0 | 0 |
| | *GARCHstat* | 1338.5 | 388.9 | 138.8 | 169.1 | 323.8 | 493.9 | 651.6 | 1019.5 | 2495.7 | 4811.8 |
| | *CriticalValue* | 3.84 | 3.84 | 3.84 | 3.84 | 3.84 | 3.84 | 3.84 | 3.84 | 3.84 | 3.84 |
| **4** | *h* | 0.76 | 0.94 | 0.96 | 0.98 | 1 | 1 | 1 | 1 | 1 | 1 |
| | *pValue* | 0.09 | 0.02 | 0.01 | 0.01 | 0 | 0 | 0 | 0 | 0 | 0 |
| | *GARCHstat* | 1031.6 | 207.1 | 293.1 | 516.1 | 683.1 | 1056.8 | 1742.9 | 2750.4 | 3717.3 | 4590.4 |
| | *CriticalValue* | 3.84 | 3.84 | 3.84 | 3.84 | 3.84 | 3.84 | 3.84 | 3.84 | 3.84 | 3.84 |

## VIII. CONCLUSION

In this paper, a novel methodology was proposed with the potential to identify and classify linear and nonlinear damages in building structures. Here, the VMD was applied to address the variational conditioning in the input signals and the GARCH model used for modeling the decomposed signals. Afterward, the IMFs were deployed as the features of input signals. It was revealed that using all IMFs led to an increase in residuals. Thus, KPCA and KDA are applied to the extracted features, respectively, to find the optimum and appropriate features. It was observed that using kernel-based dimensional reduction could enhance classification performance using SVM, KNN, and fine tree algorithms. It was demonstrated through using two empirical models that the proposed method could discriminate linear damage states correctly and without any error and classify nonlinear damages with significant accuracy. Moreover, the proposed method proves its efficiency even in a highly noisy environment with an SNR of 20 and 15. Finally, to see the existence of the GARCH effect, Kurtosis and ARCH test was deployed, and the results showed that IMFs followed the GARCH effect; thereby, they were appropriate candidates for the proposed method.

The authors suggest the application of VMD and the GARCH model for unsupervised approaches and reinforcement learning. Moreover, optimization algorithms such as particle swarm optimization (PSO) and grey wolf optimizer (GWO) could be deployed to find the optimum number of features. The current limitation of the proposed method is sensitivity to the noisy signals, which can be solved by SNR estimation and reducing the noise by signal processing approaches. Also, we can consider the semi-supervised schemes to reduces the effect of noisy features on the performance of feature reduction schemes.

## Conflict of interest

The authors declare that they have no conflict of interest.

## References

1. Nichols, J.M. and M.D. Todd, Nonlinear features for SHM applications. Encyclopedia of structural health monitoring, 2009.
2. Haroon, M., Free and Forced Vibration Models. Encyclopedia of Structural Health Monitoring.
3. Gudmundson, P., The dynamic behaviour of slender structures with cross-sectional cracks. Journal of the Mechanics and Physics of Solids, 1983. 31(4): p. 329-345.
4. Sinou, J.-J., A review of damage detection and health monitoring of mechanical systems from changes in the measurement of linear and nonlinear vibrations. 2009, Nova Science Publishers, Inc.
5. Cavadas, F., I.F. Smith, and J. Figueiras, Damage detection using data-driven methods applied to moving-load responses. Mechanical Systems and Signal Processing, 2013. 39(1-2): p. 409-425.
6. Pawar, P.M. and R. Ganguli, Structural health monitoring using genetic fuzzy systems. 2011: Springer Science & Business Media.
7. Chatzi, E.N. and C. Papadimitriou, Identification Methods for Structural Health Monitoring. Vol. 567. 2016: Springer.
8. Gopalakrishnan, S., M. Ruzzene, and S. Hanagud, Computational techniques for structural health monitoring. 2011: Springer Science & Business Media.
9. Monavari, B., SHM-based Structural Deterioration Assessment. 2019, Queensland University of Technology
10. Azimi, M., A.D. Eslamlou, and G. Pekcan, Data-Driven Structural Health Monitoring and Damage Detection through Deep Learning: State-of-the-Art Review. Sensors, 2020. 20(10): p. 2778.
11. Smarsly, K., K. Dragos, and J. Wiggenbrock. Machine learning techniques for structural health monitoring. in Proceedings of the 8th European Workshop on Structural Health Monitoring (EWSHM 2016), Bilbao, Spain. 2016.
12. Rytter, A., Variational based inspection of civil engineering structures. 1993.
13. Farrar, C.R., et al., A statistical pattern recognition paradigm for vibration-based structural health monitoring. Structural Health Monitoring, 1999. 2000: p. 764-773.
14. Gharehbaghi, V.R., et al. Deterioration and damage identification in building structures using a novel feature selection method. in Structures. Elsevier.
15. Das, S., P. Saha, and S. Patro, Vibration-based damage detection techniques used for health monitoring of structures: a review. Journal of Civil Structural Health Monitoring, 2016. 6(3): p. 477-507.
16. Khuc, T., et al., A Nonparametric Method for Identifying Structural Damage in Bridges Based on the Best-Fit Auto-Regressive Models. International Journal of Structural Stability and Dynamics, 2020. 20(10): p. 1-17.
17. Gharehbaghi, V.R., et al., Supervised damage and deterioration detection in building structures using an enhanced autoregressive time-series approach. Journal of Building Engineering, 2020. 30: p. 101292.
18. Monavari, B., et al., Structural Deterioration Localization Using Enhanced Autoregressive Time-Series Analysis. International Journal of Structural Stability and Dynamics, 2020. 20(10): p. 2042013.
19. Cheng, C., L. Yu, and L.J. Chen. Structural nonlinear damage detection based on ARMA-GARCH model. in Applied Mechanics and Materials. 2012. Trans Tech Publ.
20. Beale, C., C. Niezrecki, and M. Inalpolat, An adaptive wavelet packet denoising algorithm for enhanced active acoustic damage detection from wind turbine blades. Mechanical Systems and Signal Processing, 2020. 142: p. 106754.
21. Huang, N.E. and Z. Wu, A review on Hilbert-Huang transform: Method and its applications to geophysical studies. Reviews of geophysics, 2008. 46(2).

22. Huang, N.E., et al., The empirical mode decomposition and the Hilbert spectrum for nonlinear and non-stationary time series analysis. Proceedings of the Royal Society of London. Series A: mathematical, physical and engineering sciences, 1998. 454(1971): p. 903-995.
23. Ai, Q., et al., Advanced Rehabilitative Technology: Neural Interfaces and Devices. 2018: Academic Press.
24. Huang, B.-L. and Y. Yao, Batch-to-batch Steady State Identification via Online Ensemble Empirical Mode Decomposition and Statistical Test, in Computer Aided Chemical Engineering. 2014, Elsevier. p. 787-792.
25. Dragomiretskiy, K. and D. Zosso, Variational mode decomposition. IEEE transactions on signal processing, 2013. 62(3): p. 531-544.
26. Phan, S.K. and C. Chen, Big Data and Monitoring the Grid, in The Power Grid. 2017, Elsevier. p. 253-285.
27. Blanco-Velasco, M., B. Weng, and K.E. Barner, ECG signal denoising and baseline wander correction based on the empirical mode decomposition. Computers in biology and medicine, 2008. 38(1): p. 1-13.
28. Oladosu, G., Identifying the oil price–macroeconomy relationship: An empirical mode decomposition analysis of US data. Energy Policy, 2009. 37(12): p. 5417-5426.
29. Lee, T. and T.B. Ouarda, Prediction of climate nonstationary oscillation processes with empirical mode decomposition. Journal of Geophysical Research: Atmospheres, 2011. 116(D6).
30. Lei, Y., et al., A review on empirical mode decomposition in fault diagnosis of rotating machinery. Mechanical systems and signal processing, 2013. 35(1-2): p. 108-126.
31. Bagheri, A., O.E. Ozbulut, and D.K. Harris, Structural system identification based on variational mode decomposition. Journal of Sound and Vibration, 2018. 417: p. 182-197.
32. Maji, U. and S. Pal. Empirical mode decomposition vs. variational mode decomposition on ECG signal processing: A comparative study. in 2016 International Conference on Advances in Computing, Communications and Informatics (ICACCI). 2016. IEEE.
33. Xin, Y., J. Li, and H. Hao, Damage Detection in Initially Nonlinear Structures Based on Variational Mode Decomposition. International Journal of Structural Stability and Dynamics, 2020. 20(10): p. 2042009.
34. Das, S. and P. Saha, Performance of hybrid decomposition algorithm under heavy noise condition for health monitoring of structure.
35. Figueiredo, E., et al., Structural health monitoring algorithm comparisons using standard data sets. 2009.
36. Wang, Z., et al., Application of parameter optimized variational mode decomposition method in fault diagnosis of gearbox. IEEE Access, 2019. 7: p. 44871-44882.
37. Guo, H. and R. Zhou. Experimental research of nonlinear damage diagnosis using ARMA/GARCH method. in IOP Conference Series: Materials Science and Engineering. 2019. IOP Publishing.
38. Bollerslev, T., Generalized autoregressive conditional heteroskedasticity. Journal of econometrics, 1986. 31(3): p. 307-327.
39. Yin, S., et al. PCA and KPCA integrated Support Vector Machine for multi-fault classification. in IECON 2016-42nd Annual Conference of the IEEE Industrial Electronics Society. 2016. IEEE.
40. Yang, J., et al., KPCA plus LDA: a complete kernel Fisher discriminant framework for feature extraction and recognition. IEEE Transactions on pattern analysis and machine intelligence, 2005. 27(2): p. 230-244.
41. Schölkopf, B., A. Smola, and K.-R. Müller, Nonlinear component analysis as a kernel eigenvalue problem. Neural computation, 1998. 10(5): p. 1299-1319.
42. Cai, D., X. He, and J. Han, Speed up kernel discriminant analysis. The VLDB Journal, 2011. 20(1): p. 21-33.
43. Dukart, J., Basic Concepts of Image Classification Algorithms Applied to Study Neurodegenerative Diseases. 2015.

44. Richman, J.S., Multivariate neighborhood sample entropy: a method for data reduction and prediction of complex data. Methods in enzymology, 2011. 487: p. 397-408.
45. Tan, L., Code comment analysis for improving software quality, in The Art and Science of Analyzing Software Data. 2015, Elsevier. p. 493-517.
46. Kalbkhani, H., M.G. Shayesteh, and B. Zali-Vargahan, Robust algorithm for brain magnetic resonance image (MRI) classification based on GARCH variances series. Biomedical Signal Processing and Control, 2013. 8(6): p. 909-919.
47. Engle, R.F., Autoregressive conditional heteroscedasticity with estimates of the variance of United Kingdom inflation. Econometrica: Journal of the econometric society, 1982: p. 987-1007.
48. Park, C.-S., et al. Automatic modulation recognition of digital signals using wavelet features and SVM. in 2008 10th International Conference on Advanced Communication Technology. 2008. IEEE.